\documentclass[journal]{IEEEtran}
\usepackage{verbatim}
\usepackage[pdftex]{graphicx}
\usepackage{epsfig,subfigure}
\usepackage{amsmath,amssymb,amsfonts,bm,amscd} 
\usepackage{algpseudocode}
\usepackage{mathtools} % <-- Add this line
\usepackage{mathrsfs}
\usepackage{setspace}
\usepackage{fancyhdr}
\usepackage{algorithm} 
\usepackage{psfrag}
\usepackage{cite}
\usepackage{makecell}
\usepackage{url}
\usepackage{float}
\usepackage{soul}
\usepackage{color}
\usepackage{multirow}
\usepackage{booktabs}
\usepackage{etoolbox}

\usepackage{color, xcolor}

\newtheorem{definition}{Definition}

\newtheorem{assumption}{Assumption}
\newtheorem{remark}{Remark}

\newcommand{\mobility}[0]{ ${e}_{\text{mae}}$ ($\text{m/s}$)
                        & ${e}_{\text{max}}$ ($\text{m/s}$)     
                        & $\mathcal{E}_{\text{mae}}$ ($\text{m}$) 
                    & ${\mathcal{P}}_{d}$  ($\text{\%}$)
                    }
\newcommand{\safety}[0]
                    { Collision 
                    &  ${\mathcal{S}}_{\text{min}}$  ($\text{m}$)}
 
\newcommand{\stabilityracing}[0]{  ${\mathcal{A}}_{\text{mae}}$ ($\text{m/s}^2$) 
                       & ${\mathcal{J}}_{\text{mae}}$ ($\text{m/s}^3$)
                        & ${\mathcal{J}}_{\text{max}}$ ($\text{m/s}^3$)
                        }
\newcommand{\efficiencycruise}[0]{ ${\mathcal{A}}_{\text{mae}}$ ($\text{m/s}^2$)   
                        & ${\mathcal{J}}_{\text{mae}}$ ($\text{m/s}^3$)
                        & ${\mathcal{J}}_{\text{max}}$ ($\text{m/s}^3$)
                        }
\newcommand{\efficiencycomputation}[0]{  ${\mathcal{T}}_{solve}$ ($\text{ms}$)
                        }
\newcommand{\racingaccuracy}[0]{${\mathcal{E}}_{\text{mae}}$ ($\text{m}$) 
                    & ${\mathcal{P}}_{d}$  ($\text{\%}$)
                    & ${\mathcal{L}}_{long}$ ($\text{m}$)   
                    }
\newcommand{\rom}[1]{(\expandafter{\romannumeral #1\relax})}  
\makeatletter
\pretocmd\@bibitem{\color{black}\csname keycolor#1\endcsname}{}{\fail}
\newcommand\citecolor[1]{\@namedef{keycolor#1}{\color{blue}}}
\makeatother 

\begin{document}
\title{ 
 Spatiotemporal Receding Horizon Control with Proactive Interaction Towards Autonomous Driving in Dense Traffic 
 
} 

\author{Lei Zheng, Rui Yang, Zengqi Peng, Michael Yu Wang, \textit{Fellow, IEEE,} and Jun Ma
 \thanks{This work was supported in part by the National Natural Science Foundation of China under Grant 62303390; in part by the Guangdong Provincial Key Lab of Integrated Communication, Sensing and Computation for Ubiquitous Internet of Things under Grant 2023B1212010007; in part by the Guangzhou-HKUST(GZ) Joint Funding Scheme under Grants 2023A03J0148 and 2024A03J0618; and in part by the Project of Hetao Shenzhen-Hong Kong Science and Technology Innovation Cooperation Zone under Grant HZQB-KCZYB-2020083. \textit{(Corresponding Author: Jun Ma.)}}
  \thanks{
    Lei Zheng, Rui Yang, Zengqi Peng are with the Robotics and Autonomous Systems Thrust, The Hong Kong University of Science and Technology (Guangzhou), Guangzhou 511453, China (email: lzheng135@connect.ust.hk; ryang253@connect.hkust-gz.edu.cn; zpeng940@connect.ust.hk).}
    \thanks{Michael Yu Wang is with the School of Engineering, Great Bay University, Dongguan 523808, China (email: mywang@gbu.edu.cn).}
\thanks{Jun Ma is with the Robotics and Autonomous Systems Thrust, The Hong Kong University of Science and Technology (Guangzhou), Guangzhou 511453, China, also with the Division of Emerging Interdisciplinary Areas, The Hong Kong University of Science and Technology, Hong Kong SAR, China, and also with HKUST Shenzhen-Hong Kong Collaborative Innovation Research Institute, Futian, Shenzhen 518045, China (e-mail: jun.ma@ust.hk).} 
} 
\maketitle
  
% \markboth{IEEE Transactions on Intelligent Vehicles. Preprint Version. Accepted April, 2024}
	\maketitle

	\begin{abstract} 
     In dense traffic scenarios, ensuring safety while keeping high task performance for autonomous driving is a critical challenge. 
     To address this problem, this paper proposes a computationally-efficient spatiotemporal receding horizon control (ST-RHC) scheme to generate a safe, dynamically feasible, energy-efficient trajectory in control space, where different driving tasks in dense traffic can be achieved with high accuracy and safety in real time. 
     In particular, an embodied spatiotemporal safety barrier module considering proactive interactions is devised to mitigate the effects of inaccuracies resulting from the trajectory prediction of other vehicles. 
     Subsequently, the motion planning and control problem is formulated as a constrained nonlinear optimization problem, which favorably facilitates the effective use of off-the-shelf optimization solvers in conjunction with multiple shooting.   
     The effectiveness of the proposed ST-RHC scheme is demonstrated through comprehensive comparisons with state-of-the-art algorithms on synthetic and real-world traffic datasets under dense traffic, and the attendant outcome of superior performance in terms of accuracy, efficiency and safety is achieved.  
	\end{abstract}

      \noindent Video of the experiments: \protect\url{https://youtu.be/ezqytbIZy2A}
	\section{Introduction}
	\label{sec:introd}
    	\IEEEPARstart{A}{utonomous} vehicles (AVs) have witnessed tremendous advancements in both academia and industry over the past few decades~\cite{chen2022milestones,teng2023motion,claussmann2020review,jing2023autonomous,guo2024sustainability}. However, ensuring the safety of these vehicles remains a crucial factor in gaining widespread acceptance for their use in urban driving environments~\cite{paden2016survey,chen2021exploring, shalev2017formal,wang2023safety,TENG2024}. In dense traffic, one of the main challenges in AVs is to achieve high task performance while meeting safety requirements. It is imperative that AVs are designed to achieve both objectives effectively, providing safe and efficient autonomous driving experiences. However, interacting with other vehicles in dynamic traffic environments safely and efficiently is a significant challenge for safety-critical AVs~\cite{zhou2024interaction,gao2022accurate,brito2022learning}. For instance, during an adaptive cruise control task involving overtaking maneuvers under dense traffic conditions, the ego vehicle (EV) must maintain a stable and precise cruising speed to achieve high task performance \cite{kesting2008adaptive}. Additionally, it must return to its original lane promptly after overtaking a slower front vehicle, thereby ensuring driving consistency and safety compliance. In this highly dynamic and complex driving scenario, human drivers can exhibit unpredictable multi-modal driving behaviors, such as acceleration, deceleration, and lane changing, which are difficult to model precisely for planning and control purposes~\cite{tang2022prediction,chandra2019traphic,hu2023planning}. To avoid collisions, the EV must consider potential prediction errors of surrounding human-driven vehicles (HVs) and proactively replan its local trajectory with consideration of intricate state and input constraints~\cite{medina2024ia,claussmann2020review}. 
     % This necessitates considering both state constraints, which are vital to guarantee safe interaction with other HVs, and system dynamics, which govern the EV's movements in dense traffic. 
     However, the high complexity nature of these constraints poses a computational challenge~\cite{ ma2022alternating,huang2023decentralized}, making it difficult to plan a safe, feasible, energy-efficient trajectory to accurately accomplish its task in real time. This complexity also presents a significant challenge in balancing the tradeoff between safety and task performance when designing the objective function~\cite{gao2022accurate}. To meet these requirements, a comprehensive approach is 
     essential that considers the constraints and tradeoffs involved in the trajectory planning process. Therefore, operating in such environments requires sophisticated motion planning and control methods that take into account the uncertain intentions of other HVs, utilize the full dynamics, and ensure the safe execution of predefined tasks in real time. 
        
        In general, motion planning and control for autonomous driving applications can be attempted in a sequential manner. For the hierarchical planning and control architecture, the high-level planner plans a feasible trajectory to meet task requirements, and then a low-level feedback component executes it~\cite{paden2016survey}. However, generating a safe, feasible, and energy-efficient trajectory while taking into account full vehicle dynamics in dense traffic is computationally challenging. An informed rapidly-exploring random tree with a closed-loop controller is developed to improve sample efficiency and repair invalid reference trajectories~\cite{lin2021sampling}.  
        To realize split-second reactivity to threats, an optimized-based complimentary planner and controller sharing the same interpretation of safety for autonomous driving are proposed~\cite{wang2020infusing}. 
        However, these methods lead to high computational costs during replanning. 
        To tackle this problem, one common strategy involves decoupling the longitudinal and lateral motion to generate safe and feasible trajectories for the EV~\cite{werling2012optimal, sharath2020enhanced}. These works split the motion problem into two independent one-dimensional movements, thereby improving computational efficiency. However, it may give rise to safety concerns due to the lack of coordination between the longitudinal and lateral movements~\cite{miller2018efficient}. This becomes particularly critical when the EV needs to avoid collisions with other vehicles executing sudden lane changes in dense traffic scenarios. Alternatively, several works have employed path-velocity decomposition to simplify the motion planning problem in the path-time $(s \times t)$ space~\cite{jian2020multi,chen2022efficient}. These techniques involve planning a path to avoid static obstacles and optimizing a speed profile along the path to avoid dynamic obstacles~\cite{xu2021autonomous}. However, the reference path constrains the velocity optimization, which impacts the quality of the generated trajectory in dense traffic. To address these issues, 
        piecewise polynomials are easily optimized to meet state and control constraints to realize fast replanning for autonomous driving~\cite{han2024differential,wang2021game}. 
        % \hl{Hence, intensive works exist in optimized-based planning based on polynomial curves}.
        In these works, planners rely on the differential flatness property of the bicycle kinematic model to generate a smooth and feasible trajectory, which is further tracked by controllers considering vehicle dynamics. 
        While the generated polynomial trajectories based on the kinematic model are feasible, they cannot fully exploit the actuator potential for a nonlinear dynamic EV with nonholonomic constraints, rendering control policies sub-optimal~\cite{romero2022model}.

        Rather than addressing the planning and control problem separately, the receding horizon control (RHC) framework has been used in autonomous driving to integrate planning and control into a single joint optimization problem~\cite{liang2022novel,jin2024physical,khan2024hybrid}. RHC provides a general framework to incorporate the system constraints naturally, anticipate future events, and take control actions according to complete complex tasks encoded in the objective function. 
        However, it is non-trivial to design proper objective functions and constraints to achieve high task performance while satisfying safety requirements~\cite{borrelli2017predictive,zheng2022safe}. 
        The safety requirement is encoded as a distance term to obstacles into the hard constraints of RHC to avoid a potential collision with obstacles~\cite{karlsson2019computationally, zhang2020optimization}. As a result, the EV can generate a dynamically feasible trajectory to realize desired driving tasks based on an accurate dynamic model while avoiding collisions with obstacles. However, the reactive safety constraints represented as distance requirements in these RHC frameworks do not confine optimization until the reachable set intersects with obstacles. Nevertheless, under the reactive safety constraint setting, the EV takes no action to avoid other vehicles until they are close~\cite{zeng2021safety}, making it difficult to ensure safety in dense traffic scenarios.

        To proactively avoid collisions, a sampling-based model predictive control (MPC) approach utilizing GPU parallel sampling has been employed in~\cite{yin2022trajectory} to avoid collisions. However, it does not account for dynamic vehicles in its MPC framework. To address this limitation, a discrete nonlinear model predictive control (NMPC) based on control barrier function (CBF) has been proposed~\cite{zeng2021safety},
        allowing for successful overtaking in low-speed scenarios. 
        % In~\cite{adajania2022multi} high-speed racing
        % A rule-based adaptive MPC has been proposed to 
        Nonetheless, the optimization process in NMPC is computationally burdensome due to the need to solve the inverse of the Hessian matrix~\cite{ma2022local}. This limits the implementation for real-world autonomous driving tasks with typical optimization horizons of 5 to 10\,\text{seconds}~\cite{sadat2019jointly}.
        To facilitate efficient optimization, the alternating direction method of multipliers~\cite{boyd2011distributed} has been utilized in optimal control frameworks for autonomous driving~\cite{ma2022alternating, han2023rda}.  
        However, these works set the speed of other surrounding vehicles (SVs) to be constant. Essentially, this poses a threat to the EV's safety in dense traffic where surrounding HVs exhibit non-deterministic multi-modal behaviors.    
        Considering uncertain behaviors of surrounding HVs, researchers have extensively implemented stochastic MPC methods~\cite{brudigam2023stochastic,fu2023efficient,yin2022distributed, nair2022stochastic,bao2023moment} for the EV to avoid collisions with environmental HVs. These approaches utilize predicted probabilistic distributions of HVs' trajectories to construct safety modules based on chance constraints. Additionally, Branch MPC techniques~\cite{alsterda2021contingency, batkovic2020robust, chen2022interactive} have been adopted to account for the multi-modal behaviors of HVs by optimizing over trajectory trees.
        While most of these methods show a promising solution to enable the EV to safely interact with surrounding HVs, the optimized trajectories tend to be conservative, resulting in a comprised driving efficiency. Therefore, to overcome these rather significant impediments, a batch MPC framework has been developed to improve the computational efficiency of NMPC and address the multi-modal uncertain behaviors of other vehicles in highway scenarios~\cite{adajania2022multi}. 
        The batch MPC utilizes parallel trajectory optimization to handle the multi-modal behaviors of SVs, thus proactively avoiding collisions. While this method can generate feasible trajectories in real time based on the alternating minimization algorithm~\cite{tseng1991applications}, its trajectory evaluation algorithms could result in frequent switching between candidate trajectories, leading to aggressive driving behaviors in constrained, dense driving scenarios. 
        
       In this paper, we propose an ST-RHC framework with proactive interaction for safe and efficient autonomous driving in dense traffic scenarios. The ST-RHC framework 
       employs nonlinear programming to address the safety, dynamic feasibility, and energy efficiency in the spatiotemporal domain, enabling the EV to accomplish its driving task accurately and efficiently. Moreover, it strikes a balance between safety and task efficiency over a long prediction horizon (greater than 5\,\text{seconds}).
       To facilitate computational efficiency, we use constraint transcription and numerical parametric optimization to ensure that the nonlinear optimization problem can be solved quickly using ACADO~\cite{Houska2011a} in milliseconds.  
    
    % (TODO: can be combined together)
     The main contributions of this paper are summarized as follows:
	\begin{itemize} 
  		\item A computationally-efficient ST-RHC scheme is proposed for autonomous driving, which leverages the multiple shooting method to improve computational efficiency and numerical stability, enabling accurate accomplishment of complex tasks in dense traffic scenarios in real time.
 
    	\item A spatiotemporal safety barrier module is devised to endow the EV with proactivity and safety in dense traffic flow with uncertain HVs by designing barrier function-based safety constraints. The spatiotemporal information is utilized to mitigate the effects of prediction inaccuracies of other vehicles with uncertain driving behaviors, allowing the EV to take less conservative actions. 
	 
  		\item The improved task performance and proactive obstacle avoidance attained by our proposed framework are demonstrated  thoroughly through comparative simulations with state-of-the-art algorithms on intelligent driver model (IDM) and real-world traffic datasets. Also with high generalizability of the proposed framework, its potential can be further exploited and deployed to different driving tasks.
	\end{itemize}
      
        The rest of this paper is organized as follows. The problem statement is presented in Section~\ref{sec:problem}. The proposed methodology is described in Section~\ref{sec:alg}. The validation of the proposed algorithm applied to an autonomous vehicle system, using both synthetic and real-world traffic data, is demonstrated in Section~\ref{sec:sim}. Pertinent discussions of the results are given in Section~\ref{sec:dis}.
        Finally, a conclusion is drawn in Section~\ref{sec:con}.
	\section{Problem Statement}
	\label{sec:problem} 
    In this study, we consider a nonlinear dynamic bicycle tire model~\cite{ge2021numerically} for the ego vehicle (EV), 
	\begin{equation}
	 \Dot{\textbf{x}}(t) = f (\textbf{x}(t), \textbf{u}(t)),
	\label{eq:system_model}
	\end{equation}
	where $\textbf{x} \in \mathcal{X} \subset \mathbb{R}^{n}$ denotes the state vector, $\textbf{u} \in \mathcal{U} \subset \mathbb{R}^{m}$ denotes the control input vector, and $t$ denotes the time.  
        The state vector is defined as follows:
        \begin{equation}
        \textbf{x}=[p_x, p_y, \varphi, v_{lon}, v_{lat}, \omega]^T, 
        \end{equation}  
        where $p_x$ and $p_y$ denote the longitudinal and lateral position of the center point of the vehicle, respectively; $\varphi$ denotes the heading angle; $v_{lon}$ and $v_{lat}$ denote the longitudinal velocity and lateral velocity in the vehicle's body frame, respectively; $\omega$ denotes the yaw rate. The control input vector to the EV is defined as $\textbf{u}=[a, \delta]^T$, where $a$ and $\delta$ denote the acceleration and steering angle, respectively.

 %        The motion model for surrounding vehicles is given as:
	% \begin{equation}
	%  \Dot{\textbf{S}}(t) = f^{SV} (\textbf{S}(t)),
	% \label{eq:sv_system_model}
	% \end{equation}
 %    where $\textbf{S}={[{\textbf{S}_{p}}^T, {\textbf{S}_{v}}^T]}^T $, where $\textbf{S}_{p} = [o_{x}, o_{y}]^T $ and $\textbf{S}_{v} = [o_{v_{x}}, o_{v_{y}}]^T$ denote the position and velocity vectors of the surrounding vehicle, respectively.

        We further denote the state of the $i$-th surrounding HV as:
         \begin{equation}
         \textbf{S}^{(i)}={[{\textbf{S}^{(i)}_{p}}, {\textbf{S}^{(i)}_{v}}]}^T,
         \end{equation}
        where $\textbf{S}^{(i)}_{p} = [o^{(i)}_{x}, o^{(i)}_{y}] $ and $\textbf{S}^{(i)}_{v} = [o^{(i)}_{v_{x}}, o^{(i)}_{v_{y}}]$ represent the position and velocity vectors of the $i$-th SV, respectively.
         
    % We use the Frenet frame~\cite{Werling2010OptimalTG} as the reference frame of the centerline of the road to essentially treat curved roads as ones with a straight-line geometry. 
    With the nonlinear dynamic vehicle system (\ref{eq:system_model}) as the predictive model, an RHC scheme is designed to repeatedly solve the finite horizon optimal control problem as follows:
    \begin{alignat}{2}
    \displaystyle\operatorname*{minimize}_{\big(\textbf{x}(t),\textbf{u}(t)\big)\in\mathbb R^{n}\times \mathbb R^{m}}\ &\int_{0}^{T} \mathcal{L}(\textbf{x}(t),\textbf{u}(t))dt+ \phi(\textbf{x}(T)),\label{eq:mpc_opt1}\\
    \operatorname*{subject\ to}\quad\quad 
    &\textbf{x}(0)=\textbf{x}_0, \label{eq:mpc_opt2}\\
    &\Dot{\textbf{x}}(t)=f (\textbf{x}(t), \textbf{u}(t)),\label{eq:mpc_opt3}\\   
    &\Dot{\textbf{S}}^{(i)}(t)=\xi (\textbf{S}^{(i)}(t)) +  w^{(i)}(t),\label{eq:mpc_opt4}\\
    &\textbf{u}(t)\in \mathcal{U}, \label{eq:mpc_opt5}\\
    &\textbf{x}(t)\in \mathcal{X},	\label{eq:mpc_opt6} \\
    & \forall t\in [0, T],{\nonumber}
    \end{alignat}
     where $T$ denotes the prediction horizon; $\xi$ denotes the nominal transition model of HVs; $w^{(i)}$ denotes the motion uncertainties of the $i$-th surrounding HV. The measured initial state vector is denoted by $\textbf{x}_0$, and $\phi$ is the terminal cost function. $\mathcal{L}(\textbf{x}(t),\textbf{u}(t))$ is the running cost of the form:  
    \begin{alignat}{2}
    \small
    \mathcal{L}(\textbf{x}(t),\textbf{u}(t)) = & \|\bm{\iota}(\textbf{x}(t))\|^2_{\textbf{Q}_{1}} + \sum_{i=1}^{M} \|H(\textbf{x}(t),\textbf{S}^{(i)}(t))\|^2_{\textbf{Q}_2(\textbf{S}^{(i)}, t)}  \nonumber\\ 
    & + \|\textbf{u}(t)\|^2_{\textbf{R}},
    \label{eq:running_cost_function}
    \end{alignat}
	where $\bm{\iota}(\textbf{x})$ denotes a sparse function, which only extracts certain elements of the state vector $\textbf{x}$; $M$ denotes the number of surrounding obstacles considered in motion planning tasks; $H( \textbf{x}(t), \textbf{S}^{(i)}(t))$ denotes a continuous function for safe interaction with the $i$-th obstacles; $\textbf{Q}_1 \in \mathbb{R}^{n\times n}$, $\textbf{R} \in \mathbb{R}^{m\times m}$, and $\textbf{Q}_2(\textbf{S}^{(i)}(t), t) \in \mathbb{R}$ denote the corresponding weighting matrices, respectively. Notably, the first portion of the cost function (\ref{eq:running_cost_function}) encodes the main task for the system, e.g., racing at a desired lane or cruising at target speeds. The second term encodes safety task constraints into the requirements, e.g., obstacle avoidance. Although this term is a technically soft constraint, it provides the advantage of prioritizing safety requirements at different time steps of HVs within the prediction horizon by using varying weight coefficients $\textbf{Q}_2$. The last term regularizes the control inputs, e.g., fuel consumption and comfort of passengers. With this form of the cost function, the driving tasks can be easily encoded with a series of interpretable terms and weighted differently according to the importance of task requirements.

   In dense traffic conditions characterized by closely packed vehicles, the time headway between two vehicles is critically small. Typically, drivers maintain a time headway ranging from approximately $1\,\text{s}$ to $2\,\text{s}$ or less. Such short headways in dense traffic scenarios pose significant challenges for the EV to drive safely and efficiently. 
    To render this problem tractable, we adopt the following assumptions, akin to those presented in~\cite{shalev2017formal}:
    \begin{assumption}(\textbf{Rear-end Collision \ Liability})
    % ~\cite{shalev2017formal})
    \label{assumption: Safety_responsibility}
    In the context of vehicles driving in the same direction, if the rear vehicle collides with the front vehicle from behind, the responsibility for the accident lies with the rear vehicle. 
    \end{assumption} 
     \begin{assumption}(\textbf{Sensing \ Capabilities})\label{assumption: communication}
    An EV is equipped with perceptual sensors that enable it to gather real-time data regarding the accurate positions and velocities of nearby vehicles within its sensor range.
    \end{assumption} 
   
   The goal is to efficiently solve the problem (\ref{eq:mpc_opt1})-(\ref{eq:mpc_opt6}) in real time to accomplish specified complex tasks in  dense traffic scenarios, satisfying the following desired objectives:
    \begin{enumerate}
      \item[1)]\textit{Accuracy}: A dynamically feasible trajectory should be generated such that the task is accomplished accurately, such as enabling the EV to cruise at a target speed.   
     \item[2)]\textit{Safety}: The EV should be endowed with the ability to proactively avoid collisions with other surrounding HVs in dense traffic.
     % The EV should be endowed with the ability to mitigate the effects of prediction errors of other vehicles for proactive collision avoidance.
    \item[3)]\textit{Generalizability}: The proposed strategy should possess the versatility to be applicable to various scenarios, allowing it to track different target goals and effectively accomplish complex tasks.
    \end{enumerate}  
	\section{Methodology}
	\label{sec:alg}
    We propose a spatiotemporal RHC scheme to realize the three goals outlined in Section \ref{sec:problem}. In Section~\ref{subsec:Safety_Constraints}, 
    the spatiotemporal safety barrier module for collision avoidance with surrounding HVs is introduced. Section~\ref{subsec:Task-oriented Movement} describes the task-oriented movement for high-performance autonomous driving. The reformulation of the ST-RHC problem (\ref{eq:mpc_opt1})-(\ref{eq:mpc_opt6}) and numerical optimization are presented in Section~\ref{subsec:ST-RHC} and Section~\ref{subsection:Numerical_Optimization}, respectively.
   
	\subsection{Spatiotemporal Safety Barrier Module}
	\label{subsec:Safety_Constraints}
    In dense and dynamic driving scenarios, ensuring the safety of the EV in the presence of inaccurate trajectory predictions of other HVs should be considered appropriately in the spatiotemporal domain.  
	\subsubsection{Safety Representations}
	\label{subsec:CBF}
  \begin{figure}
		\centering
		\includegraphics[width=8cm]{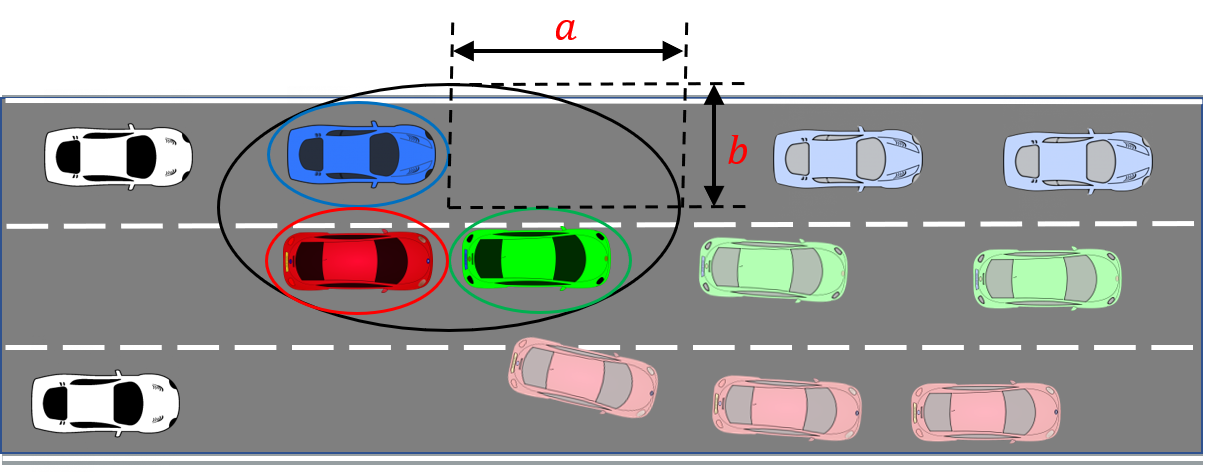}\\
		\caption{ Illustration of the two edge collision cases between the EV (shown in red) and two surrounding HVs (shown in green and blue). The shadow color represents the anticipated motion of each vehicle.} \vspace{-4mm}
  \label{fig:SafetyFunction}
	\end{figure} 
   To facilitate safe interaction between the EV and surrounding HVs, we define the $\mathit{safety}$ $\mathit{set}$ $\mathcal{S}$ of the EV system (\ref{eq:system_model}) as follows:
    \begin{equation}
    \mathcal{S}:=\{\textbf{x}(t)\in\mathcal{X}, |h(\textbf{x}(t), \textbf{S}^{(i)}(t))\geq0\}, \forall i\in\mathcal{I}_1^{M},
    \label{safety set}
    \end{equation}
    where $\mathcal{I}_1^{M}$ denote a set of integers from $1$ to $M$; $h:\mathbb{R}^{n}\rightarrow\mathbb{R}$ is a continuously differentiable barrier function that accounts for state constraints. In autonomous driving, this barrier function 
    can be represented as an ellipse that provides a safe boundary for the EV under the limiting crash case, as shown in Fig.~\ref{fig:SafetyFunction}. The barrier functions $h$ enforcing the distance between two vehicles to be larger than their safety margin can be represented as follows:  
    \begin{alignat}{2}  
   h(\textbf{x}(t), \textbf{S}^{(i)}(t)) = ||\bm{\iota}_x\textbf{x}(t) - \bm{\iota}_s\textbf{S}^{(i)}(t) ||^2_{\textbf{W}} - 1, 
    \label{eq:barrier_function}
    \end{alignat} 
    where 
    % $||\bm{\iota}_x\textbf{x} - \bm{\iota}_s\textbf{S}^{(i)} ||^2_{\textbf{W}}= (\bm{\iota}_x\textbf{x}  - \bm{\iota}_s\textbf{S}^{(i)} )^T \textbf{W} (\bm{\iota}_x\textbf{x}  - \bm{\iota}_s\textbf{S}^{(i)} )$; 
    $\bm{\iota}_x\in \mathbb{R}^{2\times 6}$, $\bm{\iota}_s\in \mathbb{R}^{2\times 4}$ are utilized to extract the longitudinal and lateral position state of the EV and $i$-th HV as follows:
    \begin{equation}
    \bm{\iota}_x = \begin{bmatrix}
    1 & 0 & 0 & 0 & 0 & 0 \\
    0 & 1 & 0 & 0 & 0 & 0
    \end{bmatrix},
    \quad
    \bm{\iota}_s = \begin{bmatrix}
    1 & 0 & 0 & 0 \\
    0 & 1 & 0 & 0
    \end{bmatrix},
    \end{equation}
   and $\textbf{W}= \mathrm{diag}( \frac{1}{a^2},\frac{1}{b^2})$ is a positive definite matrix, where $a$ and $b$ represent the length of the major and minor axes of the ellipse, respectively. 
  
  To keep the EV within a safe region $\mathcal{S}$ in dense traffic scenarios, we introduce the following definition: 
  	\begin{definition}(\textbf{Safety\ Interaction})\label{def:safety_interaction} 
            For the dynamical system~(\ref{eq:system_model}), given a continuously differentiable barrier function $h:\mathbb{R}^{n}\rightarrow\mathbb{R}$, and future trajectories of $M$ nearest HVs
            $\left\{\textbf{S}^{(i)}(t) \right\}_{t=0}^{T},\  \forall i\in\mathcal{I}_1^{M}$. The interaction between the EV and surrounding HVs is called a $\mathit{safety \ interaction}$, if there exists a control sequence $\{ \textbf{u}(t)\}_{t=0}^{T}$ in the prediction horizon $T$ such that
		\begin{equation}
		h(\textbf{x}(t), \textbf{S}^{(i)}(t))\geq0, \forall t\in\ [0,T].
		\end{equation}
	\end{definition} 
 
    \subsubsection{Spatiotemporal Safety}   
   To achieve safety interaction, holding the distance barrier function $ h(\textbf{x}) \geq 0$ is required. However, imposing this constraint as a hard requirement during planning leads to two significant drawbacks. Firstly, the constraint will not confine the optimization until the reachable set intersects with obstacles, leading to the inaction of the EV until it is quite close to other vehicles \cite{zeng2021safety}. This compromises safety and increases the risk of accidents, especially in dense traffic scenarios.
   Secondly, considering the future trajectory of HVs could result in over-conservative maneuvers, which can compromise the efficiency of the autonomous driving system. This stems from the inaccurate prediction of the future trajectories of other vehicles exhibiting multi-modal driving behaviors. This error accumulates as the prediction horizon becomes longer, leading to more significant uncertainty~\cite{li2024safe,xiong2023integrated,bao2023moment}. 
   Hence, treating all safety interactions equally throughout the prediction horizon will severely affect the generated trajectory, resulting in excessively conservative actions, such as sharply slowing down or avoiding potential interactions. This, in turn, hampers the system's ability to navigate efficiently in dynamic dense traffic conditions. 
   Note that the motion uncertainties of HVs are inherently bounded by road configurations and vehicle dynamics~\cite{huang2022survey, zhang2021exact, zhang2021bounded, guo2023vectorized}.  
 
    To address these issues, we design a spatiotemporal safety barrier module that enables the EV to proactively avoid collisions and take actions with less conservatism. With the barrier function~(\ref{eq:barrier_function}), we can design a differentiable spatial safety function $H(\textbf{x}, \textbf{S}^{(i)})$ for the safe interaction between EV and the $i$-th SV in the form of the second term in (\ref{eq:running_cost_function}) to avoid collisions as follows:
    \begin{equation} 
    \label{eq:safety_function} 
        H(\textbf{x}(t), \textbf{S}^{(i)}(t)) = \frac{1}{\lambda + h(\textbf{x}(t), \textbf{S}^{(i)}(t))} B(\textbf{x}(t), \textbf{S}^{(i)}(t)),
    \end{equation} 
    where $\lambda \in \mathbb{R^{+}}$ is a scale factor, and $B(\textbf{x}(t), \textbf{S}^{(i)}(t))$ is a safety enhancement function defined as:
    \begin{equation} 
    \label{eq:new_barrier_function} 
     B(\textbf{x}(t), \textbf{S}^{(i)}(t)) = 1 - \frac{h(\textbf{x}(t), \textbf{S}^{(i)}(t)) - c}{\eta + \sqrt{(h(\textbf{x}(t), \textbf{S}^{(i)}(t))- c)^2}} 
    \end{equation}    
    where $\eta \in\mathbb{R^{+}}$ is a small regularization constant (e.g., $\eta=10^{-5}$) that ensures numerical stability.
    For clarity, $H_i(t)$ is used to denote the safety function $H(\textbf{x}(t), \textbf{S}^{(i)}(t))$ for the $i$-th surrounding HV.
    \begin{remark} The parameter $c\in\mathbb{R^{+}}$ represents a safety measurement that adjusts the distance between two vehicles, allowing for a safety margin. This margin is critical because the value of the spatial safety function increases exponentially after crossing the safety margin, which is essential for ensuring the safety of the EV. Given reliable real-time data on SVs (Assumption 2), we maintain \(c\) as a constant to uphold a uniform safety standard in this study.  This strategy facilitates sequentially replanning the local trajectories of the EV in our ST-RHC framework with driving consistency. 
    However, a time-variant $c\in\mathbb{R^{+}}$ could be advantageous in scenarios characterized by perceptual uncertainties under rapidly changing driving conditions. This setting offers more flexible and responsive safety margins. 
    \end{remark}

 % \textbf{Temporal attention weighting matrices:}
    % Since accurately predicting the highly non-deterministic and multi-modal behaviors of vehicles is challenging, especially for long-term predictions in the presence of dense traffic flow. Therefore, overly conservative planning results can occur if safety interactions are enforced throughout the entire prediction horizon, posing a challenge to accomplishing main tasks in (\ref{eq:running_cost_function}) efficiently and accurately. 
    To address the overly conservative driving issue, we introduce a temporal attention weighting matrix $\textbf{Q}_s(\textbf{S}^{(i)}, t)$ for the spatial safety function (\ref{eq:safety_function}) as follows:
    \begin{equation}
     \textbf{Q}_s(\textbf{S}^{(i)}, t) = w_i \exp\left(\frac{-t}{\gamma}\right), i = 1,2, \cdots, M, 
    \label{eq:attention_weights_matrix}
    \end{equation}
    where $\gamma \in \mathbb{N^+}$ represents a discount factor with respect to time $t$; $w_i$ is a factor to adjust the weights for the $i$-th SV in dense traffic scenarios.  

    \begin{remark} 
    The temporal attention weighting matrix, represented by $\textbf{Q}_s(\textbf{S}^{(i)}, t)$ in (\ref{eq:attention_weights_matrix}), is a crucial component to strike a balance between the performance of the main task and the safety of the EV. This matrix gradually reduces the weight of the safety term as time passes, considering the number of SVs $M$ and the importance of the $i$-th SV. Additionally, this matrix adapts over time, reducing the weight of the spatial safety function to account for the decreasing predictability of the trajectories of other vehicles over longer prediction horizons. 
    \end{remark} 
    
    With the above spatial safety function (\ref{eq:safety_function}) and the temporal attention weighting matrix (\ref{eq:attention_weights_matrix}), a novel spatiotemporal safety barrier module is formulated as a cost for the nonlinear dynamic vehicle system (\ref{eq:system_model}) as follows:  
     \begin{equation}
    \label{eq:safety_cost}
    \small
    \begin{split}
    C_s(t) = &\sum_{i=1}^{M} \left|H(\textbf{x}(t), \textbf{S}^{(i)}(t))\textbf{Q}_s(\textbf{S}^{(i)}, t)\right|_2 \\
    = & [H_1 (t),H_2 (t), \cdots,H_M (t)]^T \\
    &\cdot diag\left(w_1, w_2, \cdots, w_M\right) \cdot \exp\left(\frac{-t}{\gamma}\right) \\
    & \cdot [H_1 (t),H_2 (t), \cdots, H_M (t)].
    \end{split}
    \end{equation} 
    The importance of safety requirements regarding different time steps in a prediction horizon can be adjusted by setting the discount factor $\gamma$ and factor $w_i$ in (\ref{eq:safety_cost}). For instance, $\gamma$ and $w_i$ can be determined according to the accuracy of trajectory prediction algorithms for different time steps and the time of prediction horizon $T$. 
  
     \begin{remark}   The spatiotemporal cost term in (\ref{eq:safety_cost}) includes a discount factor that considers the predicted time step in planning. This means the cost of a potential collision or unsafe behavior decreases as the prediction horizon increases. This balancing design between the main task and vehicle safety is achieved because the discount factor ensures that the impact of the safety cost on the overall cost function decreases over time. As a result, the planner can focus more on the main task at hand in the short term to take less conservative actions while considering the safety of the EV over the long term. 
    \end{remark} 
    
    % It is also pertinent to note that the proposed proactive approach takes into account not only the current state vector but also the future state vectors predicted based on the vehicle's dynamics and the surrounding environment. This proactivity in considering future states can lead to a safer collision avoidance maneuver as the vehicle has more time and information to plan its trajectory and make decisions.   

\subsection{Task-Oriented Movement}
 \label{subsec:Task-oriented Movement}
% Conventional motion planning and control algorithms, { such as CILQR~\cite{ma2022alternating}, }typically rely on optimization that requires a well-designed and initialized nominal trajectory. 
Our approach aims to achieve high-performance driving tasks by only considering the target goal without the requirement of a predesigned trajectory. This approach enhances the efficiency of motion planning and control and improves the overall performance of autonomous driving based on the proposed spatiotemporal safety barrier module in dense traffic flow. The task-oriented movement focuses on achieving specific driving tasks efficiently and accurately.

\subsubsection{Goal-Oriented Driving}
To accurately accomplish the desired driving tasks for the EV, it is crucial to construct goal-oriented cost terms that capture the specific requirements of each task. 
This can be achieved by designing a sparse function $\bm{\iota}_g(x)$. For instance, in the racing task with a desired driving lane, the EV should accelerate to a target velocity $v_d$ stably and rejoin its original lane after quickly overtaking its slower front vehicle. To capture these requirements, we can construct a straightforward and efficient quadratic cost term regarding the racing tasks as follows: 
\begin{equation}
     C_g = (\bm{\iota}_g(\textbf{x}-\textbf{x}_d))^T\textbf{Q}_1\bm{\iota}_g(\textbf{x}-\textbf{x}_d),
    \label{cost_cruise} 
\end{equation}
where $\bm{\iota}_g=\mathrm{diag}(0,1,0,1,0,0)$ extracts the lateral position of the vehicle and the longitudinal velocity from the state vector $\textbf{x}$, and the cost term $C_g$ serves as a constraint on the target racing lane and speed.
     \begin{figure}[tp]
		\centering
		\includegraphics[width=8cm]{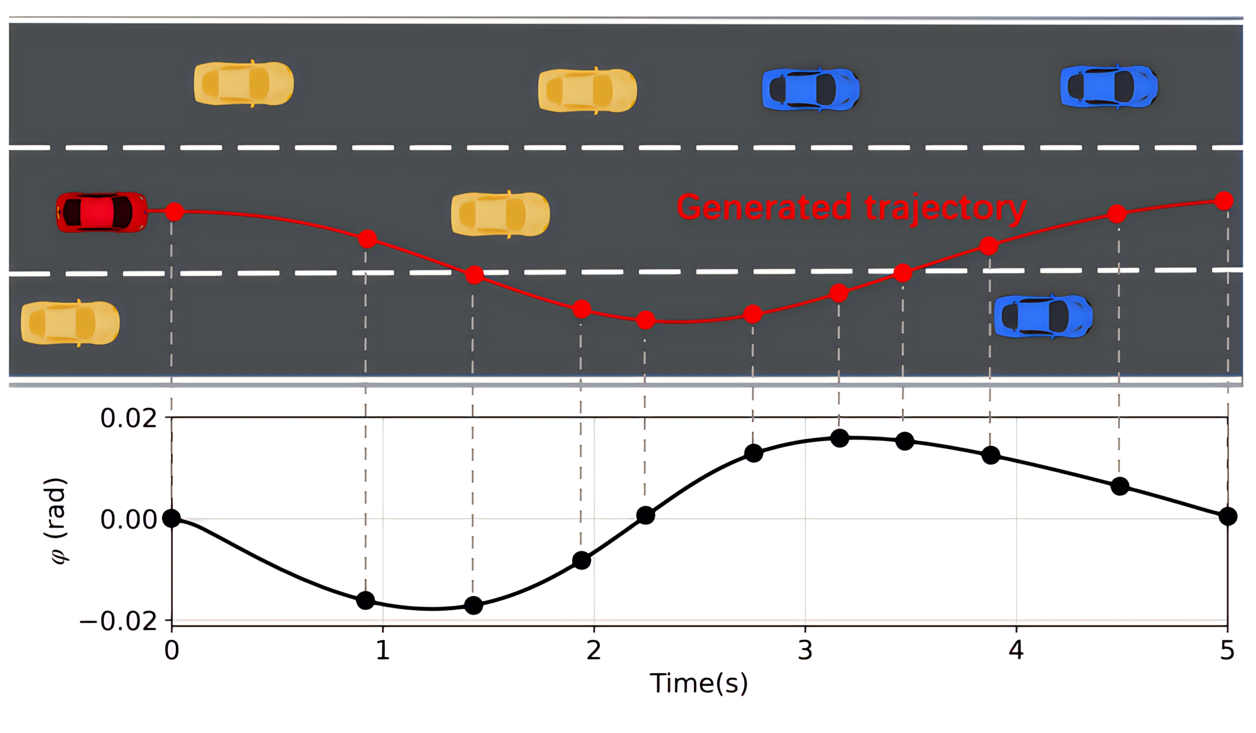}	\vspace{-2mm}
		\caption{ Top: illustration of the planned trajectory for the overtaking task, with the EV shown in red, and perceived and unperceived HVs depicted in orange and blue, respectively. Bottom: dynamic change in the heading angle $\varphi$. The heading angle decreases to a tiny value near the end of the planning horizon to stably rejoin its target racing lane. }	
	\label{fig:driving_demo}\vspace{0mm}
    \end{figure} 
\subsubsection{Driving Stability} 
\label{subsubsec:terminal_cost}
In the context of autonomous driving in dense traffic scenarios, driving stability is critical for the EV to safely and efficiently perform specific driving tasks, such as overtaking. To ensure stability, the EV should show a tiny heading angle and yaw rate to drive in its target lane at the end of the prediction horizon, thus generating a smooth trajectory as depicted in Fig.~\ref{fig:driving_demo}. 

To achieve this, the terminal cost term in the objective function (\ref{eq:mpc_opt1}) plays a crucial role in the motion planning and control process. This terminal cost should exceed the running cost to provide stability in the quasi-infinite horizon approach, as discussed in~\cite{chen1998quasi,borrelli2017predictive}. To facilitate higher stability and better driving performance of the EV, a quadratic terminal cost term $C_T$ can be designed as follows: 
\begin{eqnarray}
\left\{
    \begin{aligned}
        C_T &= \phi(\textbf{x}(t+T)) = (\bm{\iota}_T \textbf{x}(t+T))^T \textbf{Q}_T (\bm{\iota}_T \textbf{x}(t+T)), \\
        C_T &> \int_{0}^{T} \mathcal{L}(\textbf{x}(t),\textbf{u}(t)) dt,  
    \end{aligned}
\right.
\label{eq:terminal_cost}
\end{eqnarray}
where $\textbf{x}(t+T)$ is the terminal state vector, $\textbf{Q}_T$ is a large weighting matrix ensuring the terminal cost is larger than the running cost, and $\bm{\iota}_T = \mathrm{diag}(0,0,1,0,0,1)$ extracts the heading angle and yaw rate from the state 
 vector. 
% % \begin{figure}[tp]
% % 		\centering
% % 		\includegraphics[scale=0.39]{figures/mpc_sterring_angle.png}	
% % 		\caption{Top: Dynamic change in the heading angle $\varphi$ during an overtaking task with the road center-line as the reference path. The heading angle weight increases to a local maximum value at the beginning of the overtaking process, and decreases to a minimal value near the end of the planning horizon. Bottom: Visualization of the planned trajectory for the overtaking task, with the ego vehicle represented in red and the human-driven vehicle in orange.}	
% % 	\label{fig:optimization_steering_angle}
% % 	\end{figure} 
\subsection{Spatiotemporal Receding Horizon Control Scheme}
\label{subsec:ST-RHC}
 In this subsection, we reformulate the initial optimal control problem (\ref{eq:mpc_opt1})-(\ref{eq:mpc_opt6}) into a multiple shooting-based constrained nonlinear optimization problem (NLP). The primary aim of this reformulation is to leverage the benefits of direct multiple shooting methods~\cite{borrelli2017predictive} to effectively and accurately handle the complexities of nonlinear vehicle dynamics and streamline the optimization process.
 The fundamental idea of multiple shooting is to discretize the control inputs 
into $N$ shooting intervals over the horizon $T$, thereby enabling the effective resolution of the constrained NLP. Thus, we break the time domain into smaller shooting intervals as follows:  
     \begin{equation}
         0= t_0 \leq t_1 \leq  \ldots \leq t_{N-1}= T.
     \end{equation}
   By utilizing the direct multiple shooting method, we can ensure continuity between the control inputs by adding defect constraints~\cite{betts2010practical} as follows:
     \begin{equation}
     \hat{\textbf{x}}_{k} = \textbf{x}_{k+1},
    \end{equation}
    where $\hat{\textbf{x}}_{k} = \textbf{x}_{k+1}^{-} \equiv \textbf{x} ({t_{k+1}^{-}})$ 
    denotes the value of state vector $x$ at the end of the $k$-th step, while ${\textbf{x}}_{k+1}  \equiv \textbf{x} ({t_{k+1}^{+}}) = \textbf{x}_{k+1}^{+}$
    denotes the value of state vector $\textbf{x}$ at the beginning of the next step.
    
   Finally, we propose an ST-RHC scheme, considering the nonlinear vehicle dynamics (\ref{eq:system_model}) along with the goal-oriented cost $C_g$, driving stability cost $C_T$, spatiotemporal safety cost $C_s$, energy efficiency, and control limits. To efficiently solve the optimal control problem (\ref{eq:mpc_opt1})-(\ref{eq:mpc_opt6}), the ST-RHC is reformulated  as a constrained NLP as follows:
   \begin{alignat}{2}
    \small \vspace{0mm}
   \displaystyle\operatorname*{minimize}_{\big(\textbf{x}_k,\textbf{u}_k\big)\in\mathbb R^{n}\times \mathbb R^{m}}\quad
    & C_T + \sum\limits_{k=0}^{N-1} C_{g,k} + C_{s,k} + \|\textbf{u}_k\|^2_{\textbf{R}}, \label{eq:mpc_opt11}\\   \vspace{0mm}
    \operatorname*{subject\ to}\quad\quad
    &\mathrlap{0 = \textbf{x}_{k + 1} - \Phi (\textbf{x}_k, \textbf{u}_k),}\label{eq:mpc_opt12}\\
    &\mathrlap{0 = \textbf{x}_0 - \textbf{x}(0),} \label{eq:mpc_opt13}\\
    &\mathrlap{0 = \textbf{x}_{k+1} - \hat{\textbf{x}}_{k},} \label{eq:mpc_opt14}\\ 
     &\mathrlap{\textbf{S}^{(i)}_{k+1}=\hat{\xi} (\textbf{S}^{(i)}_k), \forall i\in\mathcal{I}_1^{M}, }\label{eq:mpc_opt15}\\ 
    &\mathrlap{\left[\begin{matrix} -a_{d, \max} \\ -\delta_{\max} \end{matrix}\right] \leq \textbf{u}_k \leq \left[\begin{matrix} a_{a, \max} \\ \delta_{\max} \end{matrix}\right], }\label{eq:mpc_opt16}\\
    &\mathrlap{\left[\begin{matrix} p_{x,\min} \\ p_{y,\min} \\ \varphi_{\min} \\ v_{lon, \min} \\ v_{lat, \min} \\ \omega_{\min} \end{matrix}\right] \leq \textbf{x}_k \leq \left[\begin{matrix} p_{x, \max} \\ p_{y, \max} \\ \varphi_{\max} \\ v_{lon, \max} \\ v_{lat, \max} \\ \omega_{\max}\end{matrix}\right],}\label{eq:mpc_opt17}\\
    &\mathrlap{~\forall k\in\mathcal{I}_0^{N-1},}{\nonumber}
   \end{alignat} 
   where $C_{g,k}$ and $C_{s,k}$ denote the target-oriented and spatiotemporal safety cost at the step $k$, respectively; the weighting matrix $\textbf{R}$ is a positive semi-definite diagonal matrix that determines energy efficiency; $\hat{\xi}$ represents the nominal trajectory prediction model for HVs in discrete form; $\delta_{\text{max}}$, $a_{d, \text{max}}$ and $a_{a, \text{max}}$ denote the largest steering angle, maximum deceleration, and maximum acceleration, respectively. The minimum and maximum longitudinal positions of the vehicle are represented by $p_{x,\text{min}}$ and $p_{x,\text{max}}$, respectively; $p_{y,\text{min}}$ and $p_{y,\text{max}}$ denote the lower and upper limits of the vehicle's lateral position within the road boundaries, respectively;
   $\varphi_{\text{min}}$ and $\varphi_{\text{max}}$ represent the minimum and maximum heading angles, while $v_{lon,\text{min}}$ and $v_{lon,\text{max}}$ represent the minimum and maximum longitudinal velocities. The minimum and maximum lateral velocities of the EV are represented by $v_{lat,\text{min}}$ and $v_{lat,\text{max}}$, respectively;  $\omega_{\text{min}}$ and $\omega_{\text{max}}$ denote the maximum yaw rates of the EV. The function $\Phi (\textbf{x}_k, \textbf{u}_k)$ defines the simulation of the nonlinear dynamics~(\ref{eq:system_model}) over one interval, starting from state $x_k$ and using control values $u_k$. To discretize the dynamic system~(\ref{eq:system_model}), we use 4th-order Runge-Kutta integration due to its high accuracy and numerical stability properties~\cite{butcher2016numerical}.
 
   The constraints (\ref{eq:mpc_opt12})-(\ref{eq:mpc_opt17}) facilitate the direct implementation of the control sequence $\textbf{u}^* = \{\textbf{u}(0),\textbf{u}(1), \cdots, \textbf{u}(T-1)\} $ to the dynamic vehicle system (\ref{eq:system_model}). Besides, the goal-oriented cost $C_{g,k}$, the spatiotemporal safety cost $C_{s,k}$, and the energy efficiency cost $\|\textbf{u}_k\|^2_{\textbf{R}}$ in (\ref{eq:mpc_opt11}) ensure that the EV can complete its task efficiently and safely. 
   Moreover, the driving stability can be maintained by adding an end penalty as a terminal constraint to the ST-RHC, as discussed in Section \ref{subsec:Task-oriented Movement}.
 
    The problem (\ref{eq:mpc_opt11})-(\ref{eq:mpc_opt17}) can be solved via sequential quadratic programming (SQP) to update the control sequence towards regions with lower costs. The optimal control sequence $\textbf{u}^* = \{\textbf{u}(0),\textbf{u}(1),\cdots,\textbf{u}(T-1)\} $ is obtained with the lowest cost, and the first-step control input $\textbf{u}(0)$ is applied to the vehicle system (\ref{eq:system_model}). 
  
   To ensure adaptability to dynamic changes in dense traffic, the optimal control sequence $\textbf{u}^*$ is continually updated through iteratively solving the optimization problem (\ref{eq:mpc_opt11})-(\ref{eq:mpc_opt17}) in a receding horizon manner. At each sampling instant, we use the newly measured states as the initial condition for the optimization process.  
   This optimization step involves solving a set of linear or quadratic equations to obtain a new candidate state and control sequence $\{x_{k+1}, u_k\}^{N-1}_{k=0}$. If the current optimized sequence satisfies specific convergence criteria, terminate the algorithm. Otherwise, continue to the next iteration until the algorithm reaches the maximum iteration number. 
  Such a technique offers local precision by frequently updating optimal control sequences in reaction to the evolving dynamics of nearby vehicles. Moreover, this approach of constant feedback and correction aids in promptly detecting and rectifying any deviations from the intended trajectory or behavior, thereby preserving the solution's accuracy across successive stages. 
   \textbf{Algorithm 1} details the procedures of the proposed ST-RHC.  
   
    \begin{remark}
       The spatiotemporal safety cost terms, represented by $C_s$, appear as soft constraints, but function as hard constraints as penalties are immediately imposed when the constraint boundaries are violated. Besides, the proactive level regarding collision avoidance can be adjusted by setting the prediction horizon $T$ in (\ref{eq:mpc_opt11}) and parameter $c$ in (\ref{eq:safety_function}) that governs the spatiotemporal cost term $C_s$.
    \end{remark}

\begin{algorithm}[t]
\caption{Spatiotemporal Receding Horizon Control }\label{alg:alg1}
\begin{algorithmic}[1]
\State \textbf{Parameters}: $f$: Nonlinear dynamic transition model;
\Statex \hspace{1.75cm} $\phi$, $C$: Cost function terms; 
\Statex \hspace{1.75cm} $N$: Timesteps of per roll-out; 	
\Statex \hspace{1.75cm} $\nu_0$, $\nu$: Number of optimization iterations in
\Statex \hspace{1.75cm} the initial and subsequent time steps;
\Statex \hspace{1.75cm} $M$: Numbers of the nearest HVs;
\State \textbf{initialize} the states of the $M$ nearest HVs: 
 $\left\{ \textbf{S}^{(i)}_{0} \right\}_{i=1}^{M}$;
\State \textbf{initialize} the nominal sequence $\{\Bar{\textbf{x}}_{k+1},\Bar{\textbf{u}}_k\}^{N-1}_{k=0}$;
% \State \textbf{choose} $c >0$
\State \textbf{for} $j \gets 1$ \textbf{to} $ \nu_0$ \textbf{do}:\\
 \hspace{0.5cm} Compute $\{\textbf{x}_{k+1}, \textbf{u}_k\}^{N-1}_{k=0}$ by solving the NLP 
\Statex \hspace{0.5cm} problem (\ref{eq:mpc_opt11})-(\ref{eq:mpc_opt17});
\State \hspace{0.5cm} \textbf{break} if termination criterion is satisfied;
\State \textbf{end}
\State Obtain the optimal control input $\textbf{u}^{*}=\textbf{u}_0$;
\State Send to the system~(\ref{eq:system_model}).
\State \textbf{update} $\{\Bar{\textbf{x}}_{k+1},\Bar{\textbf{u}}_k\}^{N-2}_{k=0} \gets  \{\textbf{x}_{k+1}, \textbf{u}_k\}^{N-1}_{k=1}$;
\State \textbf{while}  task is not done \textbf{do}: 
\State  \hspace{0.5cm} Measure the current state of the EV: $\textbf{x}_0$;
\State  \hspace{0.5cm} Measure the current state of $M$ nearest surrounding
\Statex \hspace{0.5cm} HVs: $\left\{ \textbf{S}^{(i)}_{0} \right\}_{i=1}^{M}$; 
\State  \hspace{0.5cm} Predict the future trajectories of the $M$ nearest 
\Statex \hspace{0.5cm} HVs: $\left\{\textbf{S}^{(i)}_{k} \right\}_{k=1}^{N},\ i = 1, .., M$;
\State \hspace{0.5cm} \textbf{for} $j \gets 1$ \textbf{to} $ \nu$ \textbf{do}:\\
 \hspace{1cm} Solve the NLP problem (\ref{eq:mpc_opt11})-(\ref{eq:mpc_opt17});
\State \hspace{1cm} \textbf{break} if convergence criterion is satisfied;
\State \hspace{0.5cm} \textbf{end} 
\State \hspace{0.5cm} Get optimal control input $\textbf{u}^{*}=\textbf{u}_0$;
\State \hspace{0.5cm} Send to the system~(\ref{eq:system_model});
\State \hspace{0.5cm} \textbf{reinitialize} the nominal state and control sequence 
\Statex \hspace{0.5cm} $\{\Bar{\textbf{x}}_{k+1},\Bar{\textbf{u}}_k\}^{N-2}_{k=0} \gets  \{\textbf{x}_{k+1}, \textbf{u}_k\}^{N-1}_{k=1}$;
\end{algorithmic}
\label{alg2}
\end{algorithm}

    \subsection{Numerical Optimization}
    \label{subsection:Numerical_Optimization}
    In this section, we analyze the characteristics of the constraints in the ST-RHC and employ numerical optimization approaches to enhance numerical stability and computational efficiency when solving this nonlinear optimization problem (\ref{eq:mpc_opt11})-(\ref{eq:mpc_opt17}).  

    The adaptive nature of the objective function (\ref{eq:mpc_opt11}) requires the EV to be able to adapt to newly detected vehicles quickly, and the ability to solve the problem within a reasonable and limited time frame, even with varying densities of HVs.  
    To tackle this challenge, we employ the SQP algorithm with a specific strategy. In the first time step, we initialize the optimization variables $\{\Bar{\textbf{u}}_k\}^{N-1}_{k=0} = \{\textbf{0}\}^{N-1}_{k=0}$ and use a relatively larger maximum iteration number $\nu_0$ than the subsequent maximum iteration number $\nu$ for optimization. 
    Subsequently, the ST-RHC framework utilizes a warm start initialization approach, where the previous control sequence $\{\textbf{u}(1),\textbf{u}(2),\cdots,\textbf{u}(T-1)\}$ is used to initialize the new control sequence (see \textbf{Algorithm 1}, lines 10 and 21). This process can reduce the number of necessary iterations to improve optimization efficiency. 
 
    The quadratic terms in the objective function (\ref{eq:mpc_opt11}) imply that using Hessian information can lead to a considerable enhancement in the speed of convergence. Nevertheless, obtaining the exact inverse Hessian matrix is computationally intensive and impractical in real-time applications. To overcome this challenge, we employ the Gauss-Newton method~\cite{gratton2007approximate} to approximate the inverse Hessian matrix from gradient information.
  
	\section{Experimental Results }
 	\vspace{0mm}
	\label{sec:sim}  
 In this section, we demonstrate the effectiveness of the proposed ST-RHC approach on a nonlinear vehicle dynamic model in the presence of uncertain multi-modal driving behaviors exhibited by human drivers, such as acceleration, deceleration, and lane changing. The high task performance and proactive obstacle avoidance capabilities of the ST-RHC in dense traffic scenarios are validated through simulations using both synthetic IDM datasets and two real-world human drivers next generation simulation\footnote{\url{https://data.transportation.gov/Automobiles/Next-Generation-Simulation-NGSIM-Vehicle-Trajector/8ect-6jqj}} (NGSIM) datasets collected from the I-80 freeway in the San Francisco Bay area.
\vspace{0mm}
   \subsection{Vehicle Model}
    The nonlinear dynamic bicycle model~\cite{ge2021numerically}  with nonlinear tire forces and simple input dynamics is formulated as follows:  \vspace{0mm}
    \begin{equation}
    \vspace{0mm}
     \Dot{\textbf{x}}=\left[\begin{array}{c}
    v_{lon}\cos \varphi-v_{lat} \sin \varphi\\
   v_{lat} \cos \varphi+v_{lon}\sin \varphi \\
    \omega\\
    a + v_{lat}\omega-\frac{1}{m}F_f sin\delta \\
    -v_{lon}\omega + \frac{1}{m}\left(F_f cos\delta + F_r \right)\\
     \frac{1}{I_z }\left(l_{f}F_f cos\delta - l_rF_r \right) 
    \end{array}\right],
    \label{eq:vehicle_dynamic_model}
    \end{equation}  
    where $m$ denotes the mass of the EV; $I_z$ denotes the polar moment of inertia; $l_f$ and $l_r$ denote the distance from the center of mass to the front and rear axles, respectively. The front and rear slip angles $\alpha_{f}$ and $\alpha_{r}$ are utilized to compute the tire lateral force of front and rear tires $F_f$ and $F_r$ as follows: 
    \begin{equation}
    \label{eq:lateral_force_front}
     F_f = k_f \alpha_f \approx k_f\left(\frac{v_{lat}+l_f\omega}{v_{lon}}-\delta\right),
    \vspace{0mm}
     \end{equation}
  \begin{equation}
    \label{eq:lateral_force_rear}
     F_r = k_r \alpha_r \approx k_r \frac{v_{lat}-l_r \omega}{v_{lon}},
\vspace{0mm}
    \end{equation}
    where $k_f$ and $k_r$ denote the cornering stiffness of the front and rear wheels, respectively. 
    % We discretize the system~(\ref{eq:vehicle_dynamic_model}) with a 4th-order Runge-Kutta integration
    %  using a sampling time of $T_s=100\,\text{ms}$ in our proposed multiple shooting ST-RHC scheme. \vspace{0mm}
     
    \subsection{Simulation Setup}
    \label{sec:setup}
    \vspace{0mm}
    We conduct the simulation experiments using 
    C++ and Robot Operating System 2 on an Ubuntu 22.04 system environment with an AMD Ryzen 5 5600G CPU with six cores @3.90 GHz with 16 GB RAM.
    We set the parameters of the nonlinear dynamic vehicle model according to \cite{ge2021numerically}, as shown in Table~\ref{table:Parameter_Settings}. We use the state-of-the-art optimization framework ACADO~\cite{Houska2011a} as the SQP solver for the constrained nonlinear optimization problem (\ref{eq:mpc_opt11})-(\ref{eq:mpc_opt17}).

  \subsubsection{ Datasets}
    In the simulation, we focus on dense traffic flow scenarios with uncertain HVs. We initially adopt the driving scenario from \cite{adajania2022multi,kesting2008adaptive} and create a six-lane unidirectional environment. At each time instance, the environment generates 18 SVs within the longitudinal range from -50 to $130\,\text{m}$ relative to the longitudinal position of the EV. 
    The minimum distance and constant safe time headway in the IDM are set to $1\,\text{m}$ and $1\,\text{m/s}$, respectively. As a result, the low-speed SVs (with speeds ranging from $7.2\,\text{m/s}$ to $12\,\text{m/s}$) result in traffic congestion, as elaborated in \cite{van2010impact}. Following the IDM, these SVs travel parallel to the centerline and adjust their cruising velocity based on the distance to the EV or the HVs ahead. The driving behavior of these SVs exhibits multi-modal characteristics, including maintaining a constant velocity, accelerating, and decelerating.  We initialize their states at the starting point of a fixed, safe lane with zero acceleration and steering angle. 
  
    To further validate the performance of our proposed strategy, we utilize two NGSIM datasets\footnote{\url{https://shorturl.at/orKR3}}: Dataset1 and Dataset2. These datasets, collected from the I-80 freeway in the San Francisco Bay area, comprise 46 and 38 HVs, respectively. {The time headway of the EV and HVs is typically from $1\,\text{s}$ to $2\,\text{s}$.}
    They showcase multi-modal driving behaviors and characteristics, such as lane changing, maintaining constant velocity, urgent acceleration, and deceleration.  
    The data in both datasets were collected at a timestep of $0.08 \,\text{s}$, ensuring a high temporal resolution for our simulations. By incorporating these diverse datasets with a fine-grained timestep into our simulation, we ensure a thorough evaluation of our proposed strategy.  
    
\begin{table}[tp]
    \centering
    \scriptsize
    \caption{General Parameters of the ST-RHC and Environmental Configurations in Driving Experiments} \vspace{-1mm}\label{table:GeneralParameter_Settings}
    \begin{tabular}{c c | c c}
    \hline
    \hline
    $\textbf{Q}_1$ & diag$(0,10^3,0,10^5,0,0)$ & $c$ & $1$ \\
    $\textbf{Q}_s$ & diag$(10^5, 10^5, 10^5, 10^5, 10^5) \times e^{-\frac{k}{5}}$ & $\eta$ & $10^{-5}$ \\
    $\textbf{R}$ & diag$(5\times10^4, 5\times10^6)$ & $\lambda$ & $1$ \\
    $\textbf{Q}_T$ & diag$(0,0,10^{10},0,0,10^8)$ & $\gamma$ & $50$ \\
    $\bm{\iota}$ & diag$(0,1,0,1,0,0)$ & $\nu_0$ & $15$ \\
    $\nu$ & $5$ & $a$ & $3\,\text{m}$ \\
    $b$ & $2\,\text{m}$ & $p_{y,\text{min}}$ & $-10\,\text{m}$ \\
    $p_{y,\text{max}}$ & $10\,\text{m}$ & $M$ & $6$ \\
    \hline
    \hline
    \end{tabular}\vspace{-2mm}
\end{table}
 
    \begin{table}[tp]
    \centering
    \scriptsize
    \caption{Parameters of Vehicle Model} \vspace{-1mm}
    \label{table:Parameter_Settings}
    \begin{tabular}{c c c}
    \hline
    Variables         & Definition  & Value   \\ \hline
    $k_f$          & Cornering
    stiffness of the front wheels & -128916 $\,\text{N/rad}$\\ 
    $k_r $          & Cornering
    stiffness of the rear wheels  & -85944  $\,\text{N/rad}$\\
    $l_f$              & Front axle distance to center of Mass & 1.06 $\,\text{m}$\\
    $l_r$              &   Rear axle distance to center of Mass & 1.85 $\,\text{m}$\\
    $m$               & Mass of vehicle  & 1412 $\,\text{kg}$\\
    $I_z$    & Polar moment of inertia  & 1536.7 $\,\text{kg} \cdot \text{m}^2$\\
    $v_{lon,\text{max}}$         & Maximum longitudinal velocity & 24 $\,\text{m/s}$\\
    $v_{lat,\text{max}}$         & Maximum lateral velocity & 3 $\,\text{m/s}$\\
    $\varphi_{\text{max}}$        & Maximum heading angle & 0.227 $\,\text{rad}$\\
    $\omega_{\text{max}}$      & Maximum yaw rate & 5 $\,\text{rad/s}$\\
    $a_{d,\text{max}}$         & Maximum deceleration & -3 $\,\text{m/s}^2$\\
    $a_{a,\text{max}}$         & Maximum acceleration & 1.5 $\,\text{m/s}^2$\\
    $\delta_{\text{max}}$      & Maximum steering angle & 0.6 $\,\text{rad}$\\
    \hline
    \end{tabular}\vspace{-2mm}
    \end{table}
 
	% \begin{table*}[tp]
	% 	\renewcommand{\arraystretch}{1.1}
	% 	\scriptsize
	% 	\caption{Performance Comparison Between Four Frameworks in Overtaking Scenario with IDM Model.}
	% 	\label{table:table_results}
	% 	\centering
	% 	\vspace{0mm}
	% 	\begin{tabular}{ c | c c  c  c c c c}
 %    	\hline
 %    		 Algorithms &  
 %            \begin{tabular}{@{}c@{}} Collision \end{tabular}  &
 %            \begin{tabular}{@{}c@{}} Cruise  MAE \\(in m/s) \end{tabular}  &
 %            \begin{tabular}{@{}c@{}} Max.  cruise \\ error (in m/s)  \end{tabular}  & 
 %            \begin{tabular}{@{}c@{}} Avg.  Lateral deviation \\ error (in m)  \end{tabular}  & 
 %            \begin{tabular}{@{}c@{}} Avg.   acceleration  \\ (in m/s$^2$)   \end{tabular} &
 %            \begin{tabular}{@{}c@{}} Min.  barrier \\ value  (in m) \end{tabular}&
 %       	\begin{tabular}{@{}c@{}} Avg. optimization \\ time (in ms) \end{tabular}\\
 %    	\hline
	% 		Batch-MPC  & No & 0.0763  &  0.4151 &0 &0.1676 &0.0230& 38.04\\			
 %                Frenet-Planner & No & 0.  &  0. &0 &0.  &0.0 & 48.04\\
	% 		RHC & No & 0.1079  &  0.5620 &0 &0.0676 &0.3260 & 16.67\\
	%        \textbf{ST-RHC} & No & \textbf{0.0176} & \textbf{0.0514}&0 & \textbf{0.0085} & \textbf{0.8174}& \textbf{12.60} \\
	% 		\hline
	% 	\end{tabular}	 
	% \end{table*}  

        \begin{table*}[t]
            \centering
            \scriptsize
            \caption{{Performance Comparison Between Four Frameworks in Cruise Scenario with IDM Dataset}}
            \label{tab:IDM_cruise_table_results}    
           \begin{tabular}[c]{|c | *{2}{c} | *{4}{c} | *{3}{c} | *{1}{c} |}
                %% HEADER
                \hline
                \multirow{2}{*}{\textbf{Algorithm}} & 
                \multicolumn{2}{c|}{{\textbf{Safety}}} &
                \multicolumn{4}{c|}{\textbf{ {\textbf{Accuracy}}}} &
                \multicolumn{3}{c|}{{\textbf{Stability}}} &
                \multicolumn{1}{c|}{{\textbf{Solving Time}}} \\
                &\safety &  \mobility  & \efficiencycruise & \efficiencycomputation    \\
                \hline
                %% ENTRIES
                        %
                \multirow{1}{*}{Frenet}
                 & No &    1.2651  &$  \textbf{ 0.0170 }$ & 0.0997 & 0.8559 & 78.5\,\% & 0.0303  & {0.0519}  & {0.3651} & 55.133    \\
                \hline           
                \multirow{1}{*}{Batch-MPC} 
                & No  &0.0230 & 0.0763  &  0.4151 & 0.8787  & 79.0\,\% &0.1676 
               & {0.1111}  & {3.6798}  & 38.04  \\
                \hline
                \multirow{1}{*}{RHC}
                  &No & 0.3260  & 0.1079  &  0.5620  & 0.9920 &  79.5\,\% & 0.0676  & \textbf{{0.0196}} & \textbf{{0.3371}}    & 16.67 \\     
                 \hline
                \multirow{1}{*}{\textbf{ST-RHC}}
                 & $ \textbf{No}$ &  0.8174 &  0.0176  & \textbf{0.0514} &  $ \textbf{0.5335}$ & $ \textbf{88.25\,\% }$ & $ \textbf{0.0085}$  & {0.0351}  & {0.9425}  & $ \textbf{12.60}$  \\
                %% END
                \hline
            \end{tabular}    \vspace{-2mm}
        \end{table*}
        \begin{table*}[t]
            \centering
            \scriptsize \vspace{-2mm}
            \caption{{Performance Comparison Between Four Frameworks in Overtaking Scenario with NGSIM Dataset}}
            \label{tab:table_results_cruise_ngsim}    
            \begin{tabular}[c]{|c | *{2}{c} | *{4}{c} | *{3}{c} | *{1}{c} |}
                %% HEADER
                \hline
                \multirow{2}{*}{\textbf{Algorithm}} & 
                \multicolumn{2}{c|}{{\textbf{Safety}}}  &  
                \multicolumn{4}{c|}{{\textbf{ Accuracy}}} &
                \multicolumn{3}{c|}{{\textbf{Stability}} }&
                \multicolumn{1}{c|}{{\textbf{Solving Time}}} \\
                &\safety &  \mobility  & \efficiencycruise & \efficiencycomputation    \\
                \hline
                %% ENTRIES
                        %
                \multirow{1}{*}{Frenet}
                 & Yes & -- & -- & - & -- &  -- & --  & --  & --  & --   \\
                \hline
                \multirow{1}{*}{Batch-MPC3} 
                 & No& 1.2855 &0.1211 & 0.5047 & 2.0369  & 50.80\,\% & 0.1143 & {1.4828}  & {34.5609}  & 53.3  \\
                \hline                
                \multirow{1}{*}{Batch-MPC6} 
                 & No& 1.3386  & 0.1448 & 0.5618  & 2.0493 & 50.40\,\%& 0.1150 &  {1.4669}   &  {17.9536} &99.1   \\
                \hline
                \multirow{1}{*}{RHC}
                  &Yes & -- & -- & - & -- &  --  & -- & -- & --  & --    \\     
                 \hline
                \multirow{1}{*}{\textbf{ST-RHC}}
                 &  $ \textbf{No}$  & 0.1941 &  $ \textbf{0.0077}$ &  $ \textbf{0.0569}$ &  $ \textbf{0.6547}$ & $ \textbf{87.60\,\%}$  & $ \textbf{0.0218}$ & \textbf{{0.0356}  }  & \textbf{ {0.4647}  }  & $ \textbf{21.01}$  \\
                %% END
                \hline
            \end{tabular}     \vspace{-2mm}
        \end{table*}
  \subsubsection{{Scenarios and Parameters}}
   At each time step, SVs are predicted to drive at a constant speed to introduce trajectory prediction errors for the EV. For NGSIM datasets, the control and communication frequency between EV and HVs are set as $12.5\,\text{Hz}$. 
    For HVs governed by the IDM, both control frequency and the communication frequency between EV and HVs are set as $10\,\text{Hz}$, with a planning time step of $T_s = 0.1\,\text{s}$. 
    We consider the following two typical driving tasks in Section \ref{overtaking_subsec} and Section \ref{racing_subsec}, respectively. 
    
    \noindent\textbf{Overtaking in adaptive  cruise scenarios}\noindent\textbf{.} The task is to keep the EV's longitudinal speed at the desired speed $v_{d}$ stably for cruising. However, if a slower vehicle is in front of the EV on the cruise lane, the EV must overtake the slower vehicle quickly and return to the cruise lane while maintaining the cruising speed as much as possible.
     The parameters of the ST-RHC and environmental settings are presented in Table~\ref{table:GeneralParameter_Settings}.   The initial acceleration and steering angle are set as $0\,\text{m/s}^2$ and $0\, \text{rad}$, respectively.
     
   \noindent\textbf{Racing in target lane scenarios}\noindent\textbf{.}  In this task, the EV is required to accelerate to a specified racing speed, denoted as $v_{d,\text{max}}$. The objective is to achieve a long travel distance with stable racing performance within a limited time. However, if a slower vehicle is ahead of the EV in the target driving lane, the EV must promptly maneuver to overtake the slower vehicle. After successfully overtaking, the EV should return to the racing lane, ensuring that it maximizes its time spent driving within the target lane. Since the racing task involves more aggressive driving maneuvers than the adaptive cruise task, we have set the terminal cost matrix $ \textbf{Q}_T = \text{diag}(0,0,10^{4},0,0,10^4)$ to allow for more flexibility in the heading angle and yaw rate changes.
     \subsubsection{{Baselines}}
    To validate the effectiveness of the proposed ST-RHC scheme, we perform an ablation study and also compare ST-RHC to the state-of-the-art baselines. 
    Specifically, comparisons with the following three approaches are performed:
	\begin{itemize}
	   \item \textbf{Batch-MPC}\cite{adajania2022multi}: A real-time multi-modal MPC algorithm designed for safe and efficient autonomous driving on highways, and we utilize its open-source codes\footnote{\url{https://github.com/vivek-uka/Batch-Opt-Highway-Driving}}  to set several parallel trajectories and tune the parameters with the best performance. The default maximum iteration number for the Batch-MPC is configured to 100, as stated in \cite{adajania2022multi}.
	   \item \textbf{Frenet planner}\cite{werling2012optimal}: A real-time  optimal sampling-based planner based on quartic polynomials for autonomous driving. We adopt the open source code\footnote{\url{https://github.com/onlytailei/CppRobotics.git}} and set four sampling horizons: 2.6, 3.6, 4.8, and 6.0\,\text{s}, with each horizon consisting of 168 trajectories along the longitudinal and lateral position of the target driving lane. 
		\item \textbf{RHC}: To demonstrate the effectiveness of the spatiotemporal safety barrier module, we conduct an ablation study by comparing the proposed ST-RHC with an ablation version, namely RHC. This version uses a fixed weighting matrix $\textbf{Q}_s = \mathrm{diag}(10^5, 10^5, 10^5, 10^5, 10^5, 10^5)$ for safety constraints.
        \end{itemize} 
  
   	\begin{figure}[tp]
		\centering
    	  \includegraphics[scale=0.35]{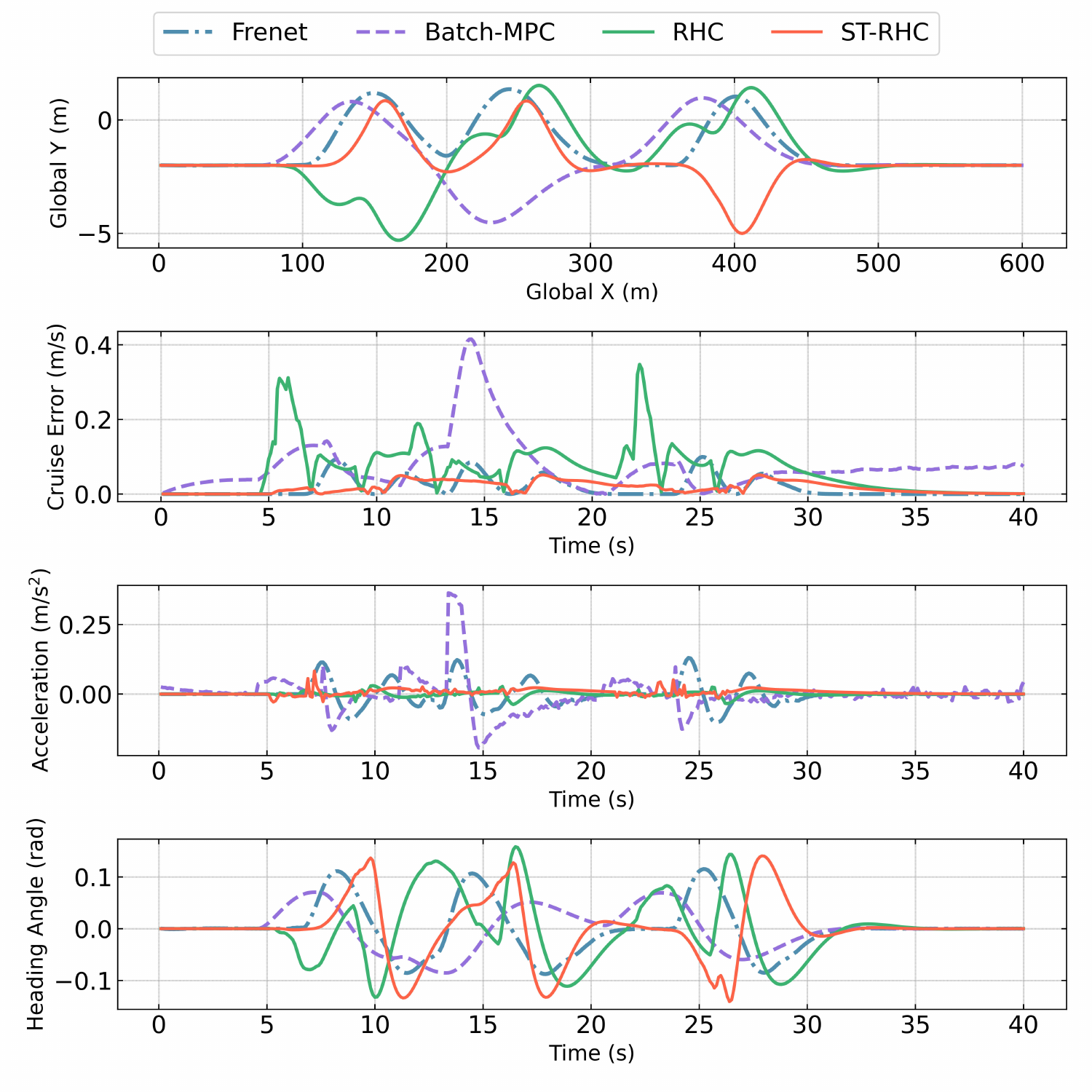}	
		\vspace{-6mm} \caption{{Comparison of position, velocity, acceleration, and heading angle profiles when executing an overtaking task using IDM dataset for surrounding HVs' motion.} The similarities in the trajectory and heading angle profiles indicate how the EV attempts to adjust its heading angle to avoid a collision with SVs.}
		\label{fig:trajs_cruise_IDM}
	\end{figure}
\subsubsection{{Evaluation Metrics}}

{Our evaluation criteria encompass four key aspects: computational efficiency, safety, task accuracy, and driving stability. Computational efficiency is measured by the average time required to solve the motion planning problem within each planning horizon. Task accuracy is assessed from various aspects, including the deviation from the target cruise speed and the lateral deviation from the intended centerline. Safety is evaluated by monitoring collisions and barrier safety values ${\mathcal{S}}_{\text{min}}$ throughout the simulation, which are calculated from the minimum barrier functions $h$. Lastly, the driving stability of the EV is measured through the analysis of the motion trajectory, acceleration, and jerk.}

	\subsection{Overtaking in Adaptive Cruise Scenarios}
	\vspace{0mm}
	\label{overtaking_subsec}
 {This subsection evaluates the performance of four algorithms using both IDM and the real-world NGSIM Dataset1\footnote{\url{https://drive.google.com/file/d/1lA3jiiNfCExrZUJIoUniFQ5i1yFQF2t0/view?usp=drive_link}}. }  
	\subsubsection{Performance Evaluation with Synthetic IDM dataset}
        \label{sussubsec:IDM_cruise}
    The simulation time, prediction horizon and time steps are set to 40\,\text{s} and $T = 5\,\text{s}$, $N =50$, respectively. The desired cruise speed and the cruise reference line are set as $15\,\text{m/s}$ and $p_{y,d} = -2\,\text{m}$, respectively.  The initial position vector of the EV is set as $[0\,\text{m},-2\,\text{m}]^T$. We set six target goal points in the target lane and the adjoint two driving lanes for the Batch-MPC. 
    
    Table~\ref{tab:IDM_cruise_table_results} summarizes the performance of the four algorithms. All four algorithms maintain positive minimum barrier function values with respect to the six nearest vehicles, ensuring the safe overtaking of slower vehicles. Notably, ST-RHC achieves the lowest maximum cruise error $e_{\text{max}}$ among the four algorithms and reduces the mean absolute error (MAE) $e_{\text{mae}}$ by $83.69\,\%$ compared to the nominal RHC. The spatiotemporal safety barrier module of ST-RHC effectively handles the uncertain behaviors of surrounding HVs, resulting in significant improvements in tracking accuracy. Furthermore, compared to Frenet, Batch-MPC and RHC, ST-RHC significantly reduces the mean absolute acceleration $\mathcal{A}_{\text{mae}}$ by $71.95\,\%$, $94.93\,\%$, and $83.69\,\%$, respectively. { This indicates that ST-RHC can generate more energy-efficiency control sequences than the other three algorithms with a relative comfort jerk value, as shown in~\cite{bae2020self}. }
    	\begin{figure}[tp]
	 	\centering  
                \subfigure[Frenet]{
			     \label{fig:IDM_Frenet-planner-demos}
			\includegraphics[scale=0.25]{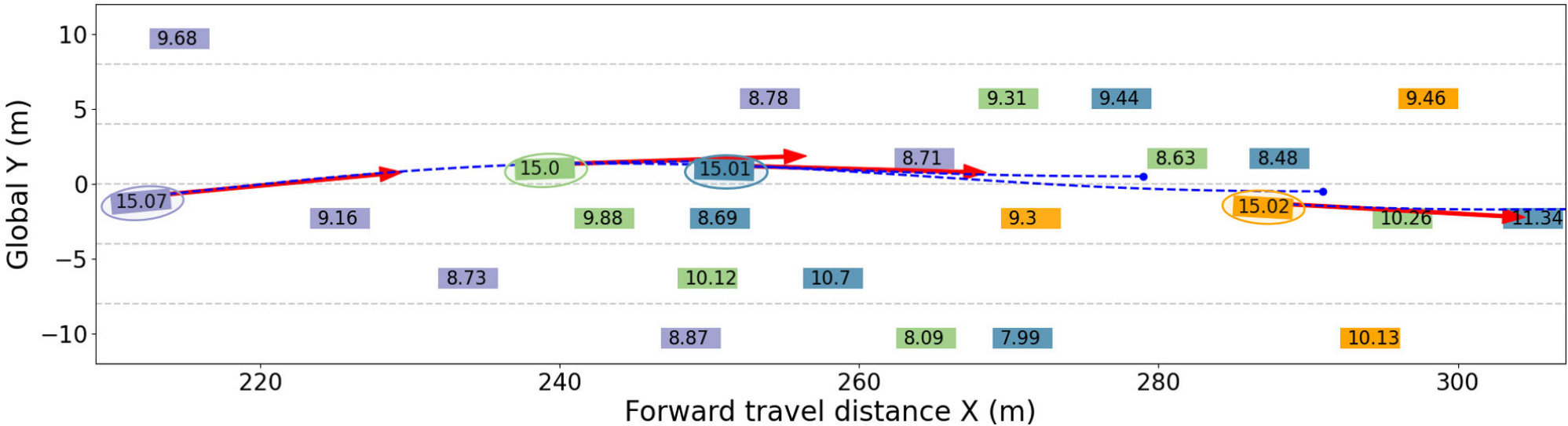}}\hspace{-2mm}
                \subfigure[Batch-MPC]{
			     \label{fig:IDM_Batch-MPC-demos}
			\includegraphics[scale=0.25]{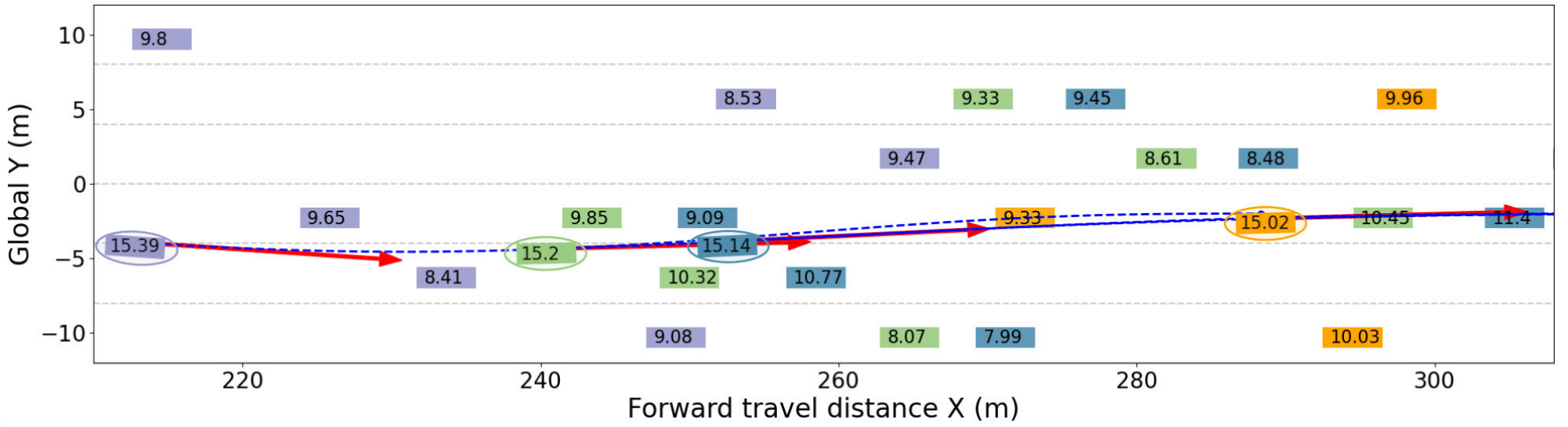}}\hspace{-2mm}
              \subfigure[RHC]{
			\label{fig:IDM_RHP-demos}
			\includegraphics[scale=0.25]{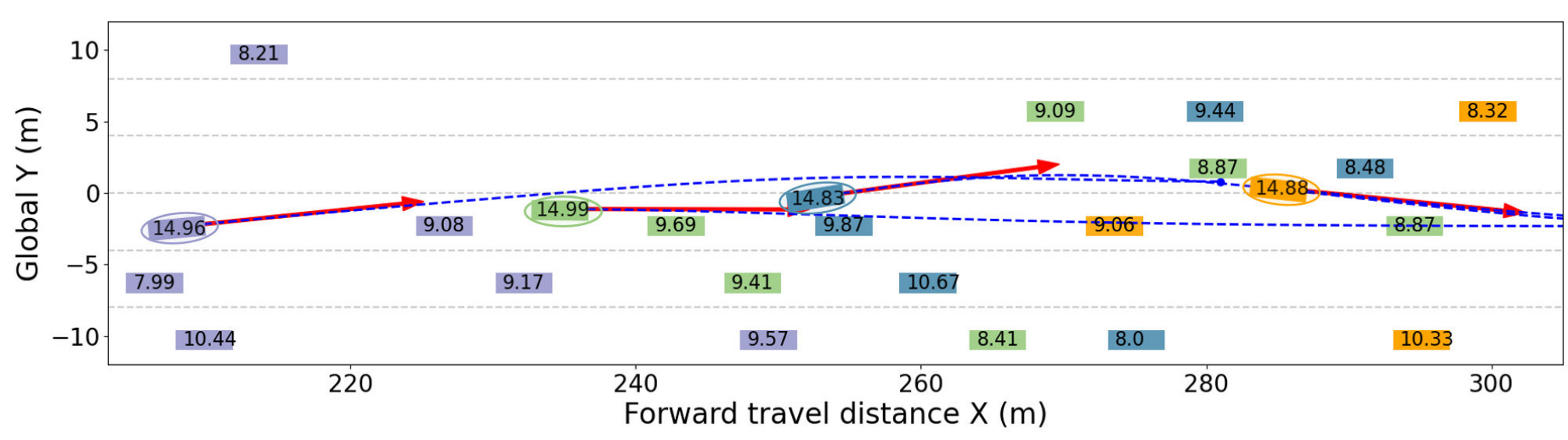}}\hspace{-2mm} 
               \subfigure[ST-RHC]{
			\label{fig:IDM_ST-RHC-demos}
			\includegraphics[scale=0.25]{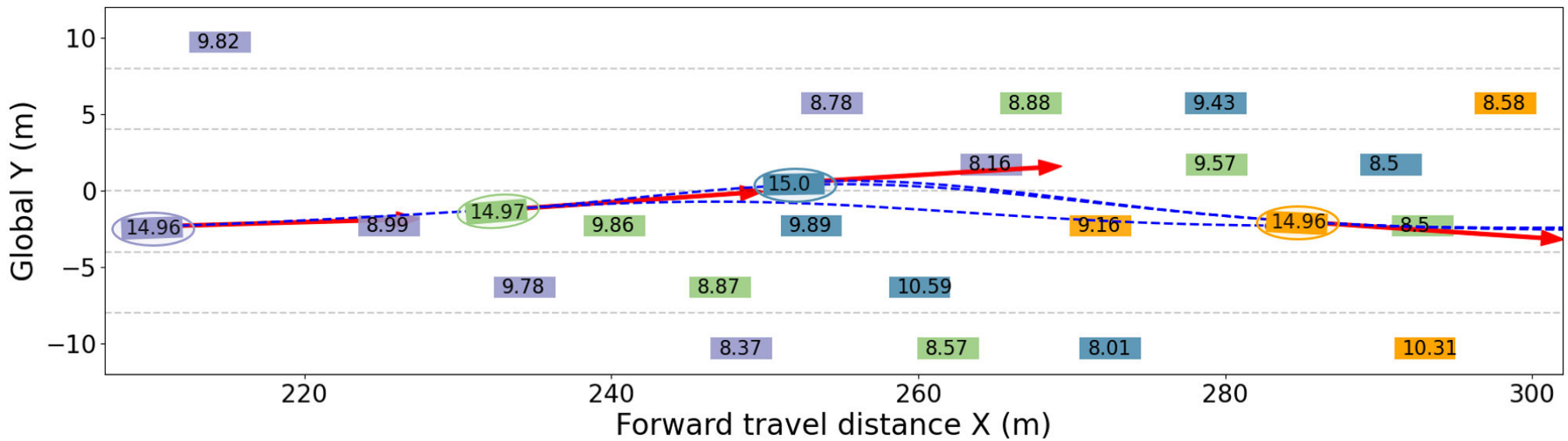}}\hspace{-2mm}
	 	\vspace{-1mm}
	 	\caption{
    Illustration of the EV's trajectory over the prediction horizon ($T = 5\,\text{s}$). The EV, represented by an ellipse, accelerates to overtake a front vehicle exhibiting multi-modal behaviors in four phases. Each phase is indicated by a unique colored rectangle, with the planned trajectory depicted as a blue dashed line. The red arrow denotes the current velocity vector of the AV. The text in each rectangle denotes the current velocity of each vehicle.}		\vspace{-1mm}
	 	\label{fig:snapots_cruise_IDM}
	 	 	\vspace{-2mm}
	\end{figure}	 
          	\begin{figure}[tp]
		\centering
		\includegraphics[scale=0.39]{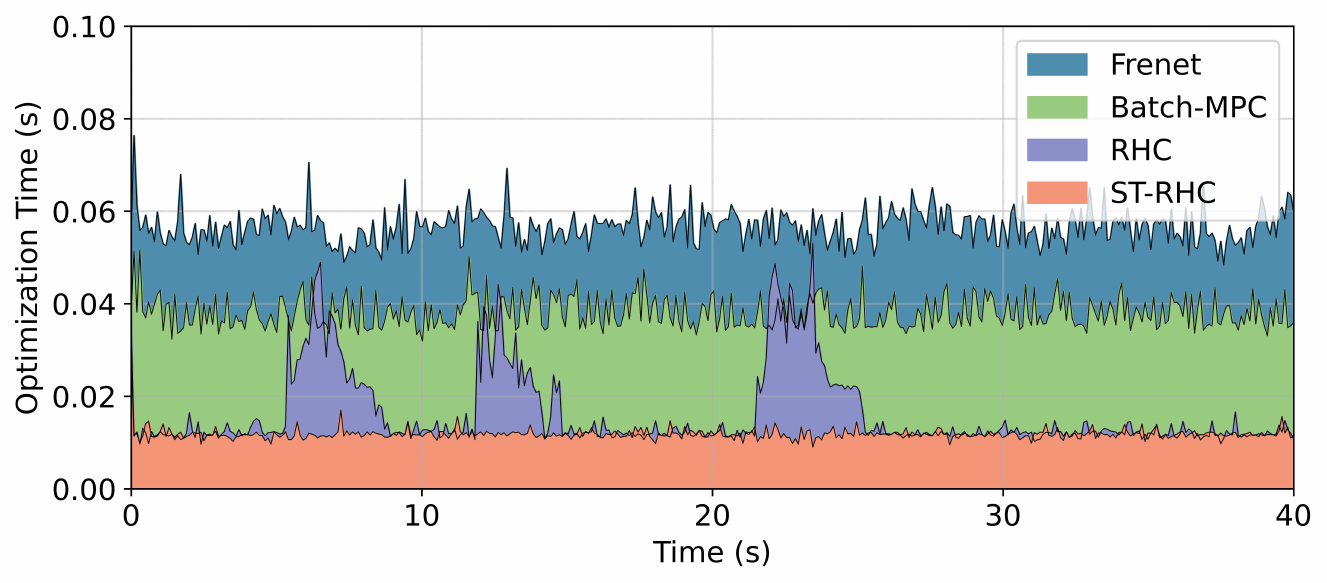}	  	\vspace{-6mm}
		\caption{Comparison of optimization time evolution for four algorithms in overtaking tasks with a prediction length of $N = 50$ in an adaptive cruise scenario, where the IDM governs the motion of surrounding HVs.}		\vspace{-2mm}
		\label{fig:optimization_time_cruise_IDM}
	\end{figure}
 
  % {The driving stability of the EV is evaluated through the evolution of motion trajectory, acceleration and jerk value, as shown in Fig.~\ref{fig:trajs_cruise_IDM}.}  

 {To better illustrate the overtaking process, Fig.~\ref{fig:snapots_cruise_IDM} shows the trajectories and velocities of the EV during overtaking maneuvers using four algorithms.} The ST-RHC, Frenet, and Batch-MPC algorithms show smoother driving maneuvers than RHC when avoiding other cars. Specifically, at $7.5\,\text{s}$, $12\,\text{s}$, and $24\,\text{s}$, the EV using ST-RHC deviates from the current cruise lane to avoid a collision with its slower dynamic vehicles in a safe manner. The results presented in Fig.~\ref{fig:trajs_cruise_IDM} and Fig.~\ref{fig:IDM_ST-RHC-demos} show that the EV can quickly rejoin the original lane with smaller acceleration after deviating from the desired lane due to obstacle avoidance. In contrast, RHC exhibits fluctuations when confronted with dynamic vehicles displaying multi-modal driving behaviors due to the accumulation of trajectory prediction errors. This is demonstrated by the trajectory and acceleration fluctuations shown in Fig.~\ref{fig:trajs_cruise_IDM}. {Consequently},  
 it necessitates a longer optimization time than the ST-RHC to generate a feasible trajectory once it becomes trapped in an unfavorable local optimization state, as depicted in Fig.~\ref{fig:IDM_RHP-demos}. This is evidenced by the three instances of fluctuating optimization time needed to avoid collisions, as shown in Fig.~\ref{fig:optimization_time_cruise_IDM}.  
As a result, The RHC exhibits a noticeably longer average optimization time $\mathcal{T}_{solve}$ than the ST-RHC ($16.67\,\text{ms}$ versus $12.60\,\text{ms}$). {These results underscore the advantages of the inherent predictive control and spatiotemporal safety barrier module in ST-RHC, which facilitate better overtaking maneuvers in cruise scenarios. This also helps to mitigate the impact of trajectory prediction errors stemming from uncertainties associated with SVs.}

Frenet and Batch-MPC, which use multiple trajectories to capture multi-modal behaviors of vehicles, require a significantly longer optimization time than ST-RHC ($55.13\,\text{ms}$ and $38.04\,\text{ms}$ versus $12.60\,\text{ms}$).  
{Furthermore, the ST-RHC algorithm outperforms Frenet and Batch-MPC by achieving the highest percentage $\mathcal{P}_{d}$ (88.25\%) of cruising within the target lane (around $-2 \pm{2} \,\text{m}$), with a minimal mean lateral deviation error $\mathcal{E}_{\text{mae}}$ from the target centerline ($-2 \,\text{m}$), resulting improved driving accuracy. This is evident from the trajectories depicted in Figs.~\ref{fig:trajs_cruise_IDM} and \ref{fig:snapots_cruise_IDM}. }
 
    \subsubsection{Performance Evaluation with Real-World Data}
      \label{sussubsec:NGSIM_cruise}
    To further show the capabilities of the proposed ST-RHC for safe and efficient interaction with uncertain HVs exhibiting muti-modal driving, we evaluate the performance of four algorithms with the NGSIM Dataset1. The simulation time, prediction horizon, and time steps are set to 20\,\text{s} and $T = 5.6\,\text{s}$, $N =70$, respectively, with a desired cruise speed of $18\,\text{m/s}$; $\textbf{Q}_1 = \mathrm{diag}(0,10^2,0,10^5,0,0)$, $p_{y,d} = 6\,\text{m}$. The initial position vector of the EV is set as $[25\,\text{m},6\,\text{m}]^T$. To achieve good planning results in this dense traffic flow with highly uncertain HVs, the max iteration number for the Batch-MPC is set as 300. To evaluate the real-time performance of Batch-MPC, we conducted experiments using two variants: Batch-MPC3 and Batch-MPC6, each representing three and six parallel optimized trajectories, respectively.

   Table \ref{tab:table_results_cruise_ngsim} shows the statistical results of five algorithms. One can notice that the Frenet and RHC algorithms are unable to maintain the EV within a safe state, mainly due to the multi-modal and highly deterministic driving behaviors exhibited by HVs. However, the Batch-MPC3, Batch-MPC6, and ST-RHC algorithms successfully facilitate the EV to handle sudden obstacles to avoid collisions with other HVs, as evidenced by the positive minimum barrier value 
  $\mathcal{S}_{\text{min}}$.  
  
    \begin{figure}
    \centering
    \includegraphics[scale=0.39]{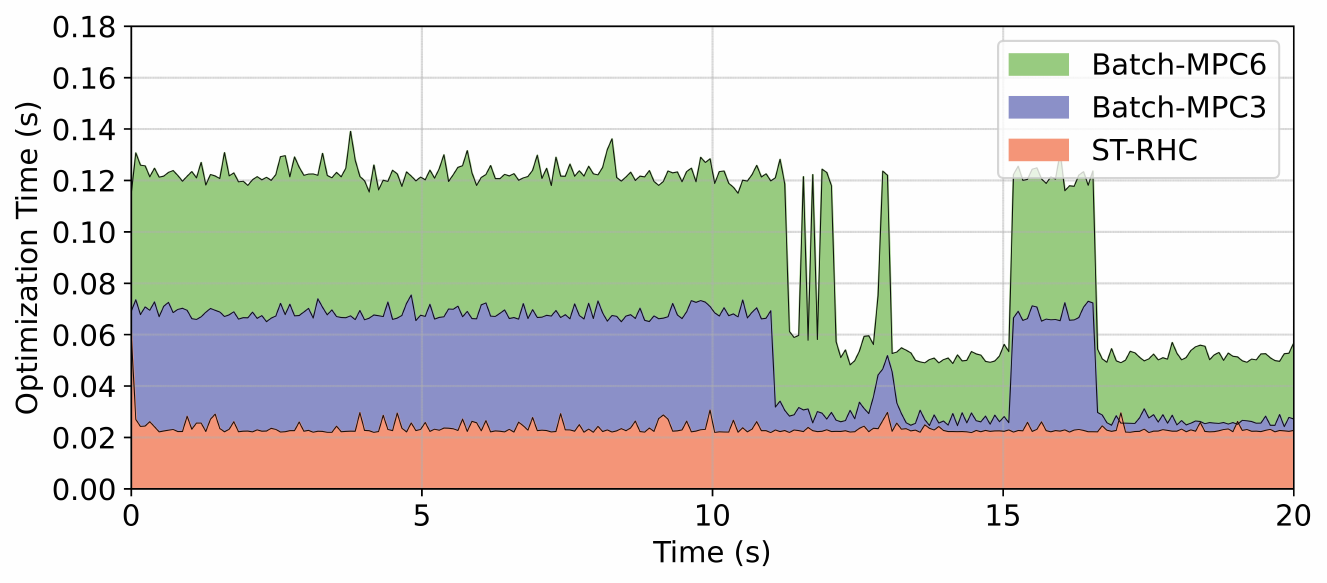}	
    \vspace{-4mm}
    \caption{Comparison of optimization time evolution for Batch-MPC and ST-RHC in overtaking tasks using a prediction length of $N = 70$ in an adaptive cruise scenario, with HVs sourced from the NGSIM dataset.}	
    \label{fig:optimization_time_cruise_NGSIM}
    \end{figure}

     \begin{figure}
    \centering
    \includegraphics[scale=0.35]{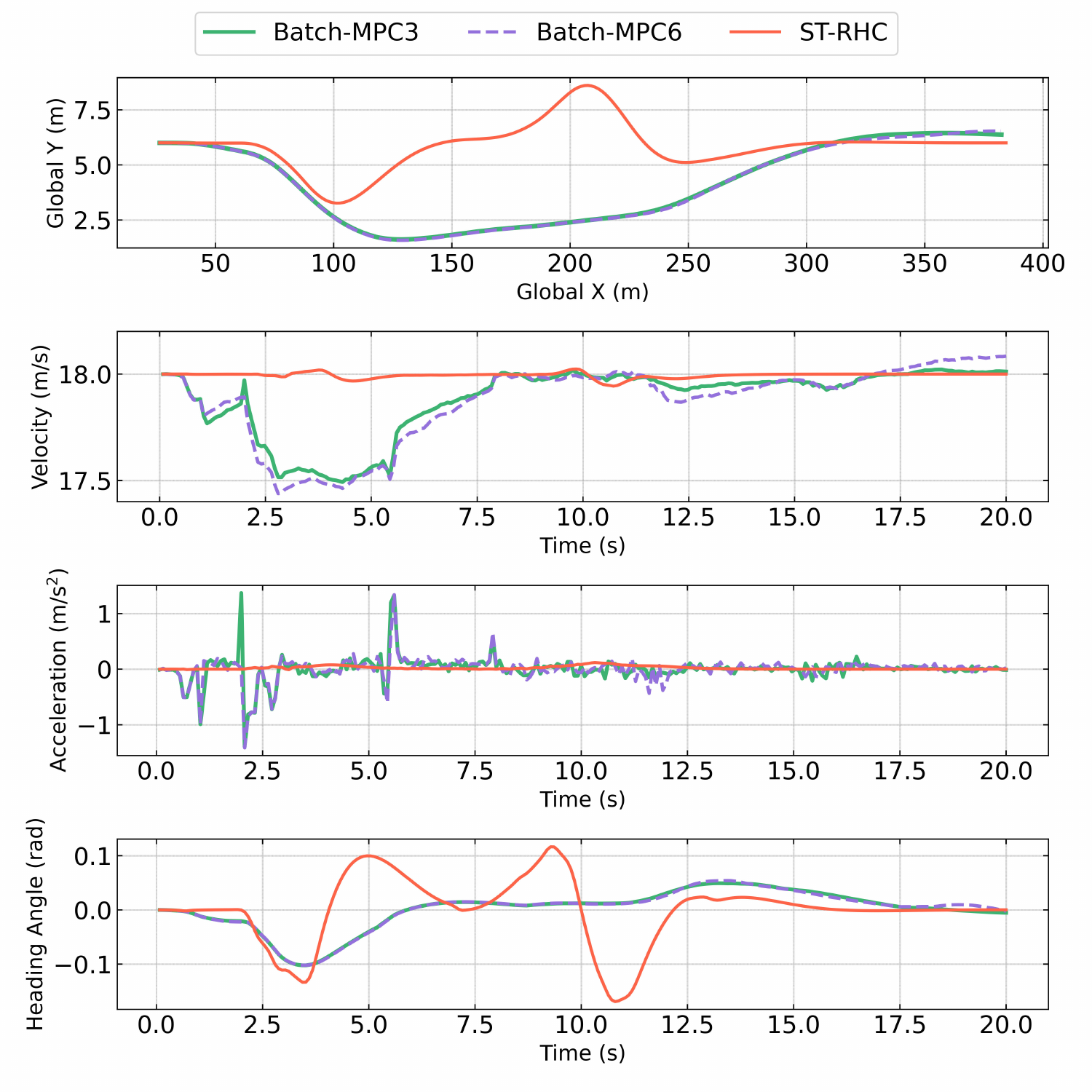}	 \vspace{-4mm} 
    \caption{Comparison of position, velocity, acceleration, and heading angle profiles based on the ST-RHC and Batch-MPC when executing an overtaking task with HVs sourced from the NGSIM dataset. The similarities in trajectory and heading angle profiles indicate that the proposed ST-RHC controller aims to adjust its trajectory to prevent collisions.}
    \vspace{0mm}
    \label{fig:performance_cruise_NGSIM}
    \end{figure}   
    In Fig.~\ref{fig:optimization_time_cruise_NGSIM}, we show the evolution of the solving time of the Batch-MPC3, Batch-MPC6, and ST-RHC. One can notice that both ST-RHC and Batch-MPC3 can achieve real-time optimization performance within a control period of $80 \text{ms}$. However,  Batch-MPC6 with six parallel optimized trajectories cannot always find a satisfactory solution with the limited computing sources under this real-world dense traffic dataset. 

     {In terms of driving stability.
     Compared to Batch-MPC3 and Batch-MPC6, ST-RHC reduces the average mean absolute acceleration value $\mathcal{A}_{mean}$ by 80.93$\,\%$ and 81.04$\,\%$, respectively. This finding is supported by the evolution of the acceleration profile depicted in Fig.~\ref{fig:performance_cruise_NGSIM}. Additionally, the largest longitudinal jerk exhibited by ST-RHC is significantly smaller than that of  Batch-MPC3 and Batch-MPC6. These observations suggest that the ST-RHC algorithm can achieve stable interactions with uncertain SVs. }
 
    {Regarding task accuracy, the ST-RHC algorithm demonstrates notably reduced cruise errors, as shown by the value of $e_{\text{mae}}$ and $e_{\text{max}}$, compared to all other algorithms. } This superiority is further supported by the evolution of velocity, as depicted in Fig.~\ref{fig:performance_cruise_NGSIM}, where the ST-RHC algorithm consistently maintains the EV closer to the desired cruise speed $v_d = 18\, \text{m/s}$. 
    Furthermore, the ST-RHC algorithm exhibits exceptional task accuracy, as evidenced by a higher percentage of time spent driving in the target lane $\mathcal{P}_{d}$. This outcome highlights its ability to effectively adhere to the desired trajectory and lane under dense and uncertain real-world traffic flow.

 \begin{table*}[t]
    \centering
    \scriptsize
    \caption{{Performance Comparison Between Four Frameworks in Racing Scenario with IDM Dataset}}
    \label{tab:table_results_racing_IDM}    
    \begin{tabular}[c]{|c | *{2}{c} |  *{3}{c} | *{3}{c} |*{1}{c} |}
        %% HEADER
        \hline
            \multirow{2}{*}{\textbf{Algorithm}} & 
        \multicolumn{2}{c|}{{\textbf{Safety}}} &
        \multicolumn{3}{c|}{\textbf{ {\textbf{Accuracy}}}} &
        \multicolumn{3}{c|}{{\textbf{Stability}}} &
        \multicolumn{1}{c|}{{\textbf{Solving Time}}}  \\ 
        &\safety & \racingaccuracy & \stabilityracing & \efficiencycomputation  \\
        \hline
        %% ENTRIES
        \multirow{1}{*}{Frenet}
         & No &   1.3241   &  $ \textbf{1.0365}$  &    $ \textbf{74.00\,\% }$  & 483.013  &  $ \textbf{0.0806}$   & {0.1414} &  {0.9685} & 55.49 \\
        \hline
        % \multirow{1}{*}{Batch-MPC3} 
        %   & Yes &  -- &  -- &  -- &  -- &   -- & -- &  -- \\
        % \hline
        \multirow{1}{*}{Batch-MPC} 
          & No &   0.0962  &  2.0883  &  37.33\,\% &   \textbf{591.588} &   0.2920  & {1.8976}  &  {20.6856} & 38.54 \\
        \hline
        %
        % \multirow{1}{*}{RHC}
        %  & Yes &  -- &  -- &  -- &  -- &   -- &  -- & -- \\       
        %  \hline
        %
        \multirow{1}{*}{\textbf{ST-RHC}}
         & $ \textbf{No}$ &  0.2186 &  1.7225  &  64.33\,\% &  586.313   & 0.1719 &\textbf{{0.0771}} & \textbf{{0.6232}}  & $ \textbf{13.09}$  \\
        %% END
        \hline
    \end{tabular}     \vspace{-2mm}
\end{table*} 

\begin{table*}[t]
    \centering
    \scriptsize
    \caption{{Performance Comparison Between Four Frameworks in Racing Scenario with NGSIM Dataset}}
    \label{tab:table_results_racing_ngsim}    
    \begin{tabular}[c]{|c | *{2}{c} |  *{3}{c} | *{3}{c} |  *{1}{c} |}
        %% HEADER
        \hline
        \multirow{2}{*}{\textbf{Algorithm}} & 
        \multicolumn{2}{c|}{{\textbf{Safety}}} &
        \multicolumn{3}{c|}{\textbf{ {\textbf{Accuracy}}}} &
        \multicolumn{3}{c|}{{\textbf{Stability}}} &
        \multicolumn{1}{c|}{{\textbf{Solving Time}}}  \\
            &\safety & \racingaccuracy & \stabilityracing & \efficiencycomputation \\
        \hline
        %% ENTRIES
        \multirow{1}{*}{Frenet}
         & Yes &  -- &  -- &  -- &  -- &  & -- &  -- & -- \\
        \hline
        % \multirow{1}{*}{Batch-MPC3} 
        %   & Yes &  -- &  -- &  -- &  -- &   -- &  -- & -- \\
        % \hline
        \multirow{1}{*}{Batch-MPC3} 
          & Yes &  -- &  -- &  -- &  --   &  -- & -- & -- &  -- \\
        \hline
       \multirow{1}{*}{Batch-MPC6} 
          & Yes &  -- &  -- &  -- &  --  &  -- & -- & -- &  -- \\
        \hline
        %
        % \multirow{1}{*}{RHC}
        %  & Yes &  -- &  -- &  -- &  -- &   -- &   -- & -- \\       
        %  \hline
        %
        \multirow{1}{*}{\textbf{ST-RHC}}
         & $ \textbf{No}$ &  0.1690 &  $ \textbf{1.0632}$ & $ \textbf{76.71\,\%}$& $ \textbf{497.751 }$ &$ \textbf{0.1548}$  & \textbf{{0.029}} & \textbf{{1.350}}  & $ \textbf{24.04}$ \\
        %% END
        \hline
    \end{tabular}     \vspace{-2mm}
\end{table*}
\begin{figure}[tp]
    \centering
      \includegraphics[scale=0.35]{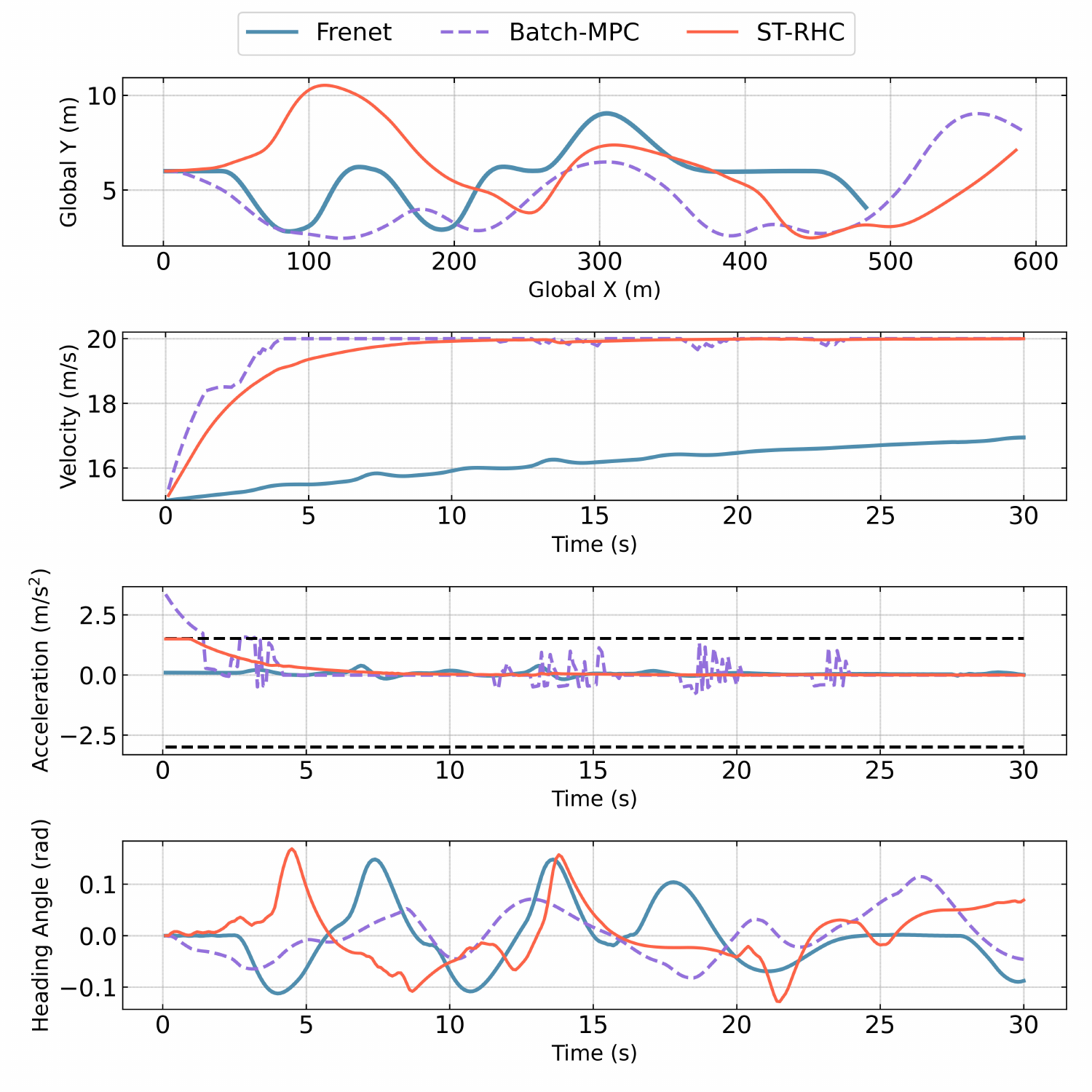}   \vspace{-2mm}
    \caption{{Comparison of position, velocity, acceleration, and heading angle profiles when executing a racing task using a prediction length of $N = 50$, where the IDM controls the SVs}.} 
    \label{fig:trajs_racing_IDM}
\end{figure} 
    \subsection{Racing in Target Lane Scenarios}
    \label{racing_subsec}
    In this subsection, {we further validate the performance of our proposed algorithm ST-RHC in a challenging, aggressive high-speed racing scenario. As the RHC fails to safely accomplish this challenging task, its results are not presented here.}  
    \subsubsection{Performance Evaluation with Synthetic IDM dataset}   
        \label{sussubsec:IDM_racing}
  We set the racing scenario with a desired speed of $20\,\text{m/s}$ and a cruise centerline at $p_{y,d} = 6\, \text{m}$. The simulation lasts for 30\,\text{s}, and the initial state vector of the EV is $[0\,\text{m}, 6\,\text{m}]^T$.  All other settings remain the same as those in the overtaking task in the adaptive cruise control scenarios, as described in Section~\ref{sussubsec:IDM_cruise}.
    
    Table~\ref{tab:table_results_racing_IDM} presents the performance comparison of the Frenet, Batch-MPC, and ST-RHC in the racing scenario. {The Frenet shows a high percentage of racing in the desired racing lane ($\mathcal{P}_{d} = 74\,\%$).}  However, its travel efficiency, represented by $\mathcal{L}_{long}$, is significantly lower compared to Batch-MPC and ST-RHC ($483.013\,\text{m}$ versus $591.588\,\text{m}$ and $586.313\,\text{m}$). This observation is further supported by the evolution of racing trajectories and velocity depicted in Fig.~\ref{fig:trajs_racing_IDM}. Additionally, ST-RHC outperforms Batch-MPC in driving accuracy in the desired lane ($\mathcal{P}_{d}= 74\,\%$ versus  $37.33\,\%$) and energy efficiency represented by $\mathcal{A}_{\text{mae}}$  with similar forward travel distance ($586.31\,\text{m}$ versus $591.59\,\text{m}$ ). 
    
  \begin{figure}[tp]
    \centering
    \includegraphics[scale=0.39]{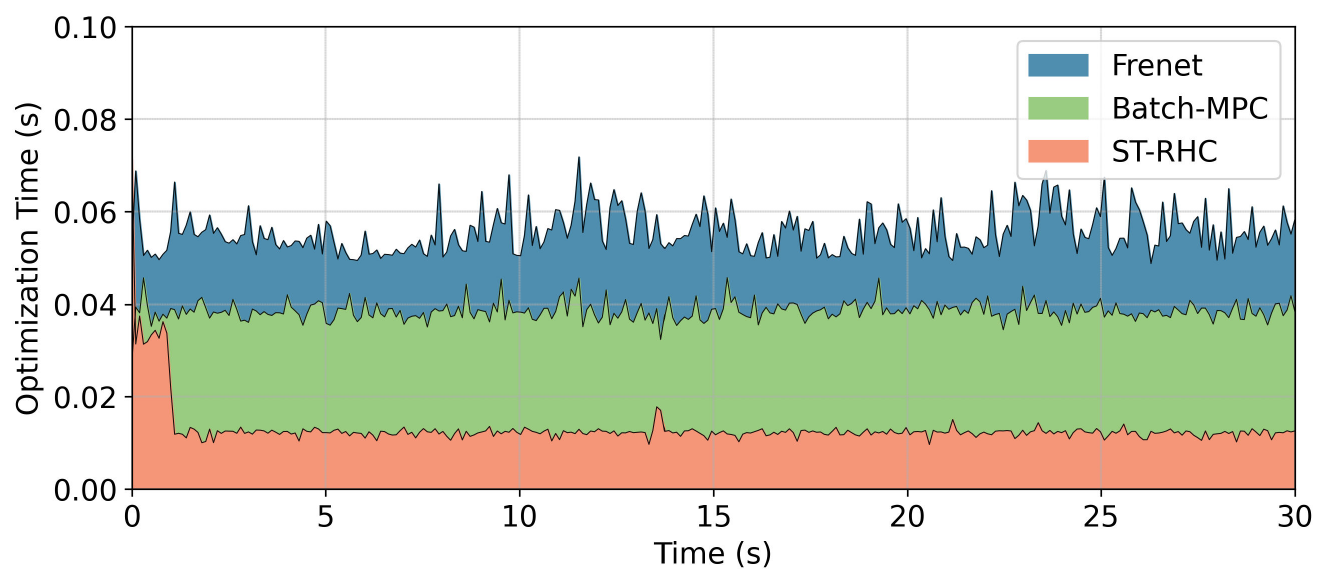}     \vspace{-0mm}
    \caption{Comparison of optimization time evolution for different algorithms when executing a racing task using a prediction length of $N = 50$, where the IDM controls the SVs.}	 
    \label{fig:optimization_time_racing_IDM}
\end{figure}
    {Regarding racing stability,} {compared to Batch-MPC, ST-RHC reduces the mean absolute jerk $\mathcal{J}_{\text{mae}}$ and maximum jerk $\mathcal{J}_{\text{max}}$ by $95.94\,\%$ and $96.99\,\%$, respectively. These observations showcase that the ST-RHC algorithm can achieve stable driving performance in this demanding racing task.} Additionally, it is important to note that Batch-MPC violates the maximum acceleration limit ($a_{a, \text{max}} = 1.5\,\text{m/s}^2$), as illustrated in the acceleration subfigure in Fig. \ref{fig:trajs_racing_IDM}. This can be attributed to Batch-MPC finding locally optimal solutions within a given maximum iteration number to ensure real-time performance, as shown in Fig. \ref{fig:optimization_time_racing_IDM}. 

    \subsubsection{Performance Evaluation with Real-world Data} 
   {To further showcase the capabilities of ST-RHC in achieving high-task performance while safely interacting with uncertain HVs in dense traffic, this subsection evaluates the performance of four algorithms using the NGSIM Dataset2\footnote{\url{https://drive.google.com/file/d/1mDUdDlt5VLifFaQthJMcANC0wrvDcqAO/view?usp=drive_link}}.} The simulation time is set to 34\,\text{s}; $\textbf{Q}_1 = \text{diag}(0, 50, 0, 10^3, 0, 0)$; $p_{y,d} = -6\, \text{m}$. The initial position vector of the EV is set as $[-10\,\text{m},-6\,\text{m}]^T$. All other settings are consistent with those used in the racing scenarios, with a maximum iteration number of 300 for Batch-MPC, as described in Section~\ref{sussubsec:IDM_racing}.
    
	\begin{figure}[tp]
		\centering
    	\hspace{-3mm}\includegraphics[scale=0.2]{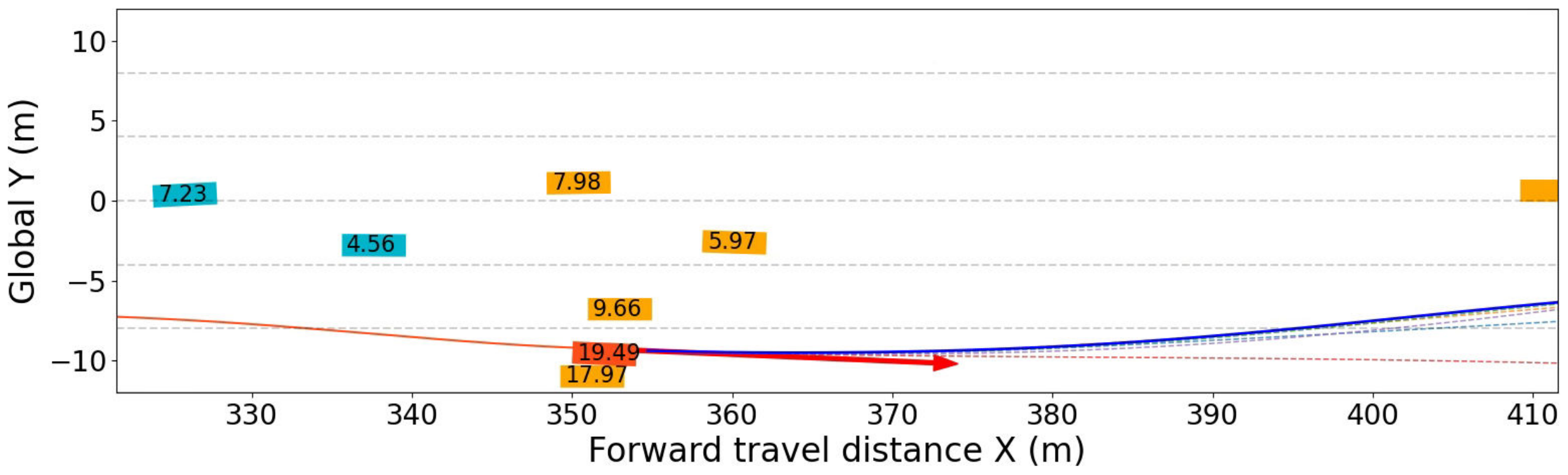}	
		\caption{ Snapotshot of the collision over the prediction length of $N = 70$ in the racing scenarios based on the Batch-MPC at 18.8 \text{s}. } 
		\label{fig:snapots_racing_NGSIM}
	\end{figure} 
    
   	\begin{figure}[tp]
		\centering
    	\hspace{-2.5mm}\includegraphics[scale=0.34]{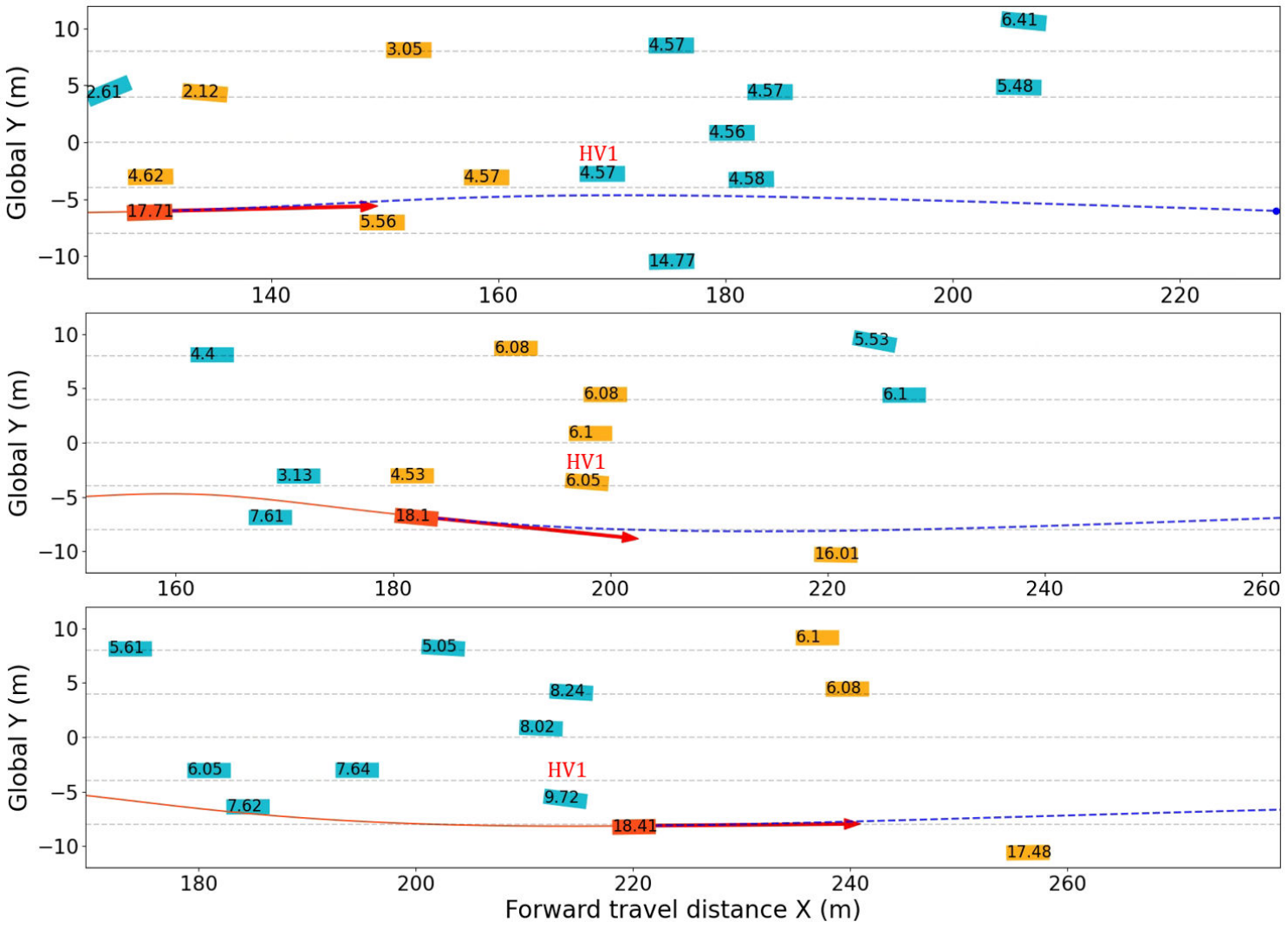}	
		\vspace{0mm}
		\caption{ Snapshots of the EV's trajectory over the prediction length ($N = 70$) in the racing scenarios at 8.48\, \text{s}, 11.6\, \text{s}, 13.52\, \text{s} based on ST-RHC. The EV, depicted as a red rectangle, accelerates to surpass front vehicles exhibiting multi-modal behaviors.}  
		\label{fig:snapots_racing_NGSIM_strhc}
	\end{figure}
 
    Table~\ref{tab:table_results_racing_ngsim} reveals that both the Frenet and Batch-MPC algorithms are unable to enable the EV's safe racing through the challenging dense traffic flow. This observation indicates that the accumulated trajectory prediction error of other multi-modal HVs significantly hinders their performance in this demanding high-speed racing task.
    In Fig.~\ref{fig:snapots_racing_NGSIM}, an illustration highlights a collision between the red EV and an orange human-driven vehicle using Batch-MPC, even after employing a large iteration number of 300. This occurrence can be attributed to the following underlying reasons:
    
    (i) The pre-recorded HVs from the NGSIM dataset exhibit multi-modal driving behaviors and are unable to react to the presence of the EV, leading to a failure in avoiding collisions.
    
    (ii) The constant velocity motion prediction model used for the surrounding HVs resulted in the accumulation of large prediction errors with a long prediction length $N=70$, which became especially problematic in high-speed racing tasks under a dense traffic flow.
 
    (iii)  The motion of the non-holonomic EV is restricted by control limits when interacting with highly uncertain HVs that have larger acceleration and steering angle bounds in the dense traffic scenario. This constraint further exacerbated the likelihood of a collision, as the EV's maneuverability and ability to avoid obstacles are restricted by its non-holonomic constraints and the unpredictable and aggressive driving behaviors exhibited by the surrounding HVs.
    
\begin{figure}[tp]
    \centering
       \includegraphics[scale=0.385]{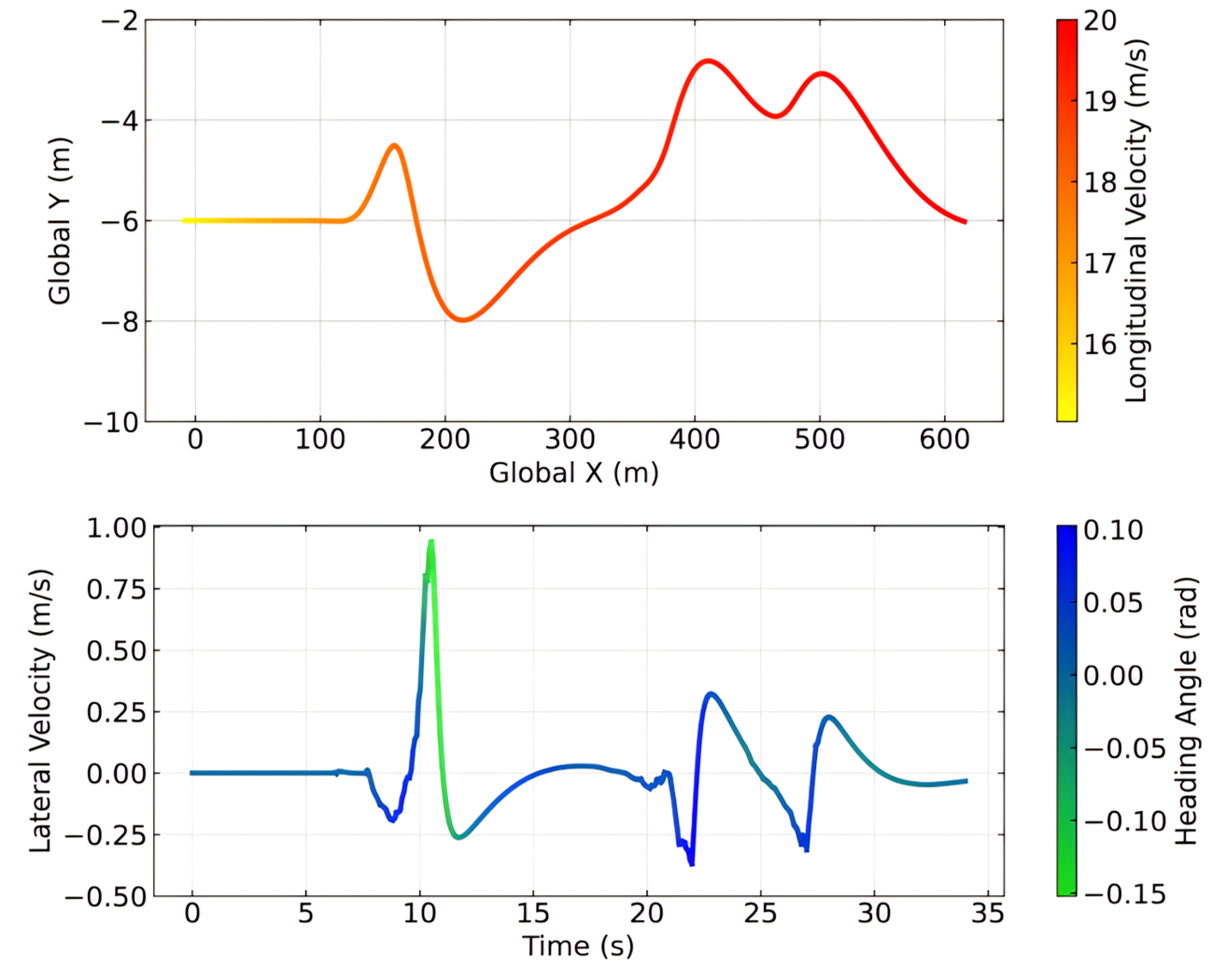}
    \caption{Driving trajectory and velocity in a racing scenario with a target speed of  $20\,\text{m/s}$  based on the ST-RHC. The top subfigure displays the trajectories of the EV, colored according to their speed profile (yellow-red color). The bottom subfigure shows the lateral velocity of the EV, colored according to their heading angle profile (green-blue color).}	
        \label{fig:colored_racing_NGSIM}
        \vspace{0mm}
    \end{figure}

    On the other hand, as illustrated in Fig.~\ref{fig:snapots_racing_NGSIM_strhc}, 
   an imperceptible blue HV1 starts executing a lane change to its right lane at 8.48\, \text{s}. The EV detects the HV1 and promptly adjusts its driving direction and planned trajectory towards the bottom lane at 11.6 $\text{s}$ to avoid a potential collision with the HV1 undergoing a lane change into the EV's current lane. This highlights the effectiveness of the ST-RHC in facilitating the EV's safe navigation through dense traffic.
    This is further substantiated by its positive barrier safety value ${\mathcal{S}}_{\text{min}}$.
    Fig.~\ref{fig:colored_racing_NGSIM} provides visualizations of the racing trajectories and lateral velocity of the EV. Notably, the EV efficiently adjusts its lateral velocity to avoid collisions. Besides, it reduces the lateral velocity and absolute heading angle value to rejoin the target driving lane after avoiding surrounding HVs. These results show that the ST-RHC can account for the multi-modal behaviors of HVs and proactively change its state to interact with other human drivers in dense traffic safely.

	\begin{figure}[tp]
		\centering
		\includegraphics[scale=0.39]{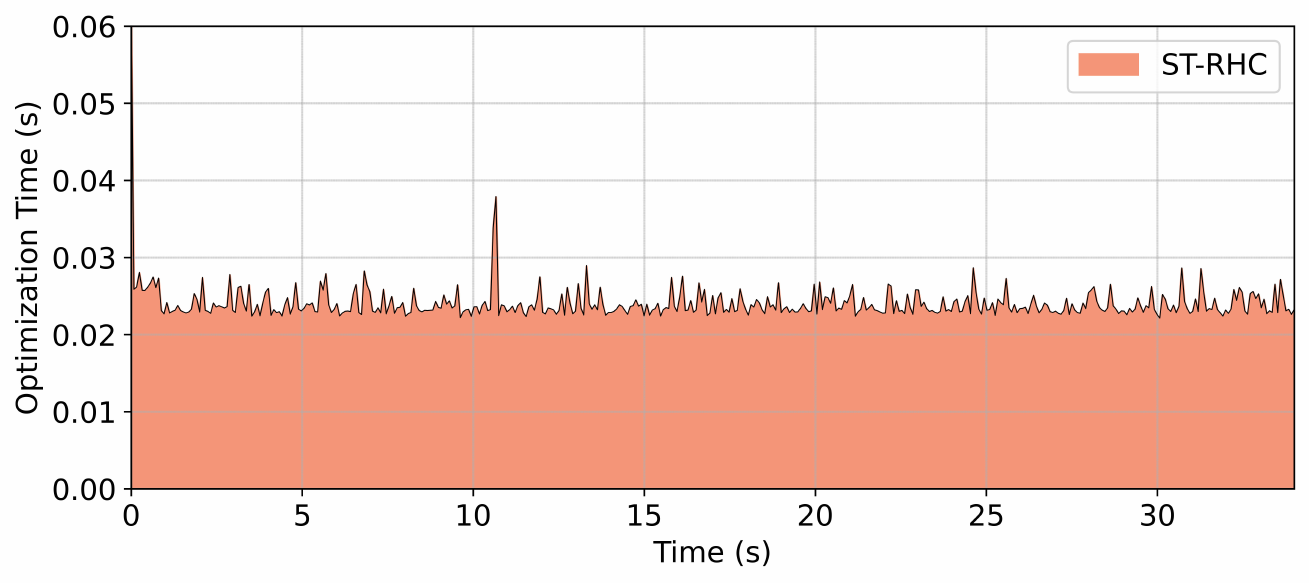}	
	    \vspace{-0mm}
		\caption{ Evolution of optimization of  ST-RHC with prediction steps $N=70$ in the racing scenario in NGSIM Dataset1.}	\label{fig:optimization_time_raing_NGSIM}  
	\end{figure} 
    Regarding computational efficiency, the average solving time of the ST-RHC $\mathcal{T}_{\text{solver}}$ is $24.04\,\text{ms}$ with the prediction length $N =70$. This is further supported by the evolution of optimization time, as depicted in Fig.~\ref{fig:optimization_time_raing_NGSIM}. 
     
    Overall, these simulation results demonstrate the effectiveness of the ST-RHC approach in achieving high task performance and safety under challenging dense traffic flow, utilizing both IDM and real-world traffic datasets. Notably, these superior outcomes are achieved with real-time execution with long optimization horizons ( $T \geq 5\,\text{s}$ and $N \geq 50$). 
    
    \section{Discussions}
	\label{sec:dis} 
\subsection{Real-Time Performance Analysis}
 
\begin{table}[t]
    \centering
    \scriptsize
    \caption{Average Optimization Time Comparison with Different Numbers of HVs within Adaptive Cruise Scenario.}
    \label{tab:computational_time}
    \begin{tabular}{ccccccc} 
        \toprule
        Number of Considered Obstacles & \multicolumn{4}{c}{Average Optimization Time (in ms)} \\
        \cmidrule(r){2-5}
         & ST-RHC & Frenet & Batch-MPC \\
        \midrule
        {4} &  \textbf{9.76} & 51.20 & 25.44 \\
          {6} & \textbf{10.55 }& 51.81  &  32.58 \\
          {8} & \textbf{10.8} &  53.10 &  42.28 \\
          {10} & \textbf{10.92} & 54.19 &  51.88\\
        \bottomrule
    \end{tabular} 
\end{table}
In this subsection, 
{we investigate the effects of varying collision avoidance constraints on the computation time within the ST-RHC framework. } Our objective is to assess how the algorithm's computational efficiency adapts to the different number of safety constraints in dense traffic scenarios. 
{For this analysis, we adopt an adaptive cruise control scenario with a prediction horizon of  $N = 50$ and aim for a replanning interval of 100\,\text{ms}.} 
To systematically explore the impact of collision avoidance constraints, we conducted experiments with a progressively increasing number of anticipated nearest HVs.  
Specifically, we configured 24 HVs, strategically distributed over a longitudinal range from -50\,\text{m} to 130\,\text{m} relative to the position of the EV. 

As depicted in Table~\ref{tab:computational_time}, a linear increase is observed in the average optimization time concerning the number of obstacles for Batch-MPC. 
In contrast, one can notice that an increase in the number of collision avoidance constraints leads to a discernible impact on the average computation time for the Frenet and ST-RHC. The increase in the number of obstacles does not impact the sampling process for the Frenet algorithm, which accounts for the majority of time in the trajectory planning process. Consequently, this increase does not substantially affect the overall time consumption.   
Regarding ST-RHC, the optimization process maintains an average time of approximately  10\,\text{ms} for ST-RHC, thereby supporting a control frequency surpassing 50\,\text{Hz}. This outcome indicates that adding more obstacles does not significantly complicate the optimization problem, showcasing the algorithm's robust scalability and effectiveness in handling dense traffic scenarios. 
It also underscores our commitment to ensuring the algorithm remains practical and meets real-time requirements. 
\begin{table*}[tp]
		\renewcommand{\arraystretch}{1.1}
		\scriptsize
		\caption{Speed Tracking Control and Safety Performance With Different Parameters and Configurations}
		\label{table:table_predictivity} \vspace{0mm}
		\centering
		\begin{tabular}{cc|ccccccc}
			\hline
      		\begin{tabular}{@{}c@{}} Target Cruise \\  Speed (in m/s) \end{tabular}  & 
			\begin{tabular}{@{}c@{}} Prediction\\Horizon  (in $\text{s}$)\end{tabular}  &  
			\begin{tabular}{@{}c@{}} Distance to avoid 1st \\ dynamic vehicle (in $\text{m}$)\end{tabular}  & 
			\begin{tabular}{@{}c@{}} Min. barrier \\ value (in $\text{m}$)\end{tabular}  &
			\begin{tabular}{@{}c@{}} Max. tracking \\  error (in m/s) \end{tabular}& 
			\begin{tabular}{@{}c@{}}  MAE \\ (in m/s) \end{tabular}& 			
   \begin{tabular}{@{}c@{}} Mean absolute \\  Acc. (in $\text{m/s}^2$)   \end{tabular}& 
   \begin{tabular}{@{}c@{}} Max. absolute\\ Acc. (in $\text{m/s}^2$)   \end{tabular}& 
   			\begin{tabular}{@{}c@{}} Avg. optimization \\ time (in ms) \end{tabular}\\
			\hline
			10  & 2   &  2.514 &  0.5130 & 0.4774 & 1.064$\times 10^{-2}$ & 1.2286$\times 10^{-2}$ &0.4332 & \textbf{2.11} \\
                10  & 5  &4.748 & 0.2592 & 0.0740 & 2.956$\times 10^{-3}$ & 0.5630$\times 10^{-2}$&0.3422  & 13.03\\
                10  & 8  &  \textbf{5.796} & 0.2588 & \textbf{0.0639} & \textbf{2.687$\times 10^{-3}$} & \textbf{0.5426}$\times 10^{-2}$ &\textbf{0.1393} & 40.69 \\
                \hline
			12 & 2  & 4.604 & 0.0874  & 0.2216 &6.744$\times 10^{-3}$ & 1.1167$\times 10^{-2}$&0.5028 & \textbf{2.20}\\
			  12 & 5  & 5.897 & 0.3790 & 0.0647 & 3.928$\times 10^{-3}$ & \textbf{0.7513}$\times 10^{-2}$ &0.2725 & 13.75\\
			12 & 8  & \textbf{7.312} &   0.5510   &  \textbf{0.0584} &\textbf{ 2.941$\times 10^{-3}$}& 0.8257$\times 10^{-2}$ &\textbf{0.2413}& 42.24\\
   	    \hline
			15  & 2   & 9.925 & 0.5432 & 0.1146&  7.812$\times 10^{-3}$& 2.5092$\times 10^{-2}$& 0.5920&\textbf{2.17}\\
			15  &5   &  13.661 & 0.6354 & 0.0242& \textbf{4.957$\times 10^{-3}$}& \textbf{1.1757}$\times 10^{-2}$ &0.0739 & 13.26\\	
			15  & 8   & \textbf{16.894} &   0.9805  & \textbf{0.0408} & 5.570$\times 10^{-3}$ & 1.4868$\times 10^{-2}$ & \textbf{0.0724} &40.83\\
			\hline
		\end{tabular} \vspace{0mm}
	\end{table*}  
 
    \subsection{Tradeoff Between Safety and Task Performance}
    \label{tradeoff between safety and stability} 
    % \textbf{Tradeoff Between Safety and Task Performance:}
    To assess the tradeoff between safety and task performance, as well as the generalizability of our proposed strategy for tracking different speeds, we conducted simulations using ST-RHC with various target cruise speeds $v_d$ and prediction horizons $T$ in an adaptive cruise overtaking scenario, similar to the one in Section~\ref{sussubsec:IDM_cruise}. Specifically, we test our approach for target cruise speeds of $v_d = 10\,\text{m/s}$, $v_d = 12\,\text{m/s}$, and $v_d = 15\,\text{m/s}$, with simulation durations of $65\,\text{s}$, $50\,\text{s}$, and $20\,\text{s}$, respectively. 
  
    Simulation results are listed in Table~\ref{table:table_predictivity}. We highlight four key results: computational efficiency, safety performance, cruise error, and energy efficiency.
    As for the computational efficiency, it can be seen that the average optimization time is less than $100\,\text{ms}$ in each task, indicating that the optimization can be performed in real time for each configuration. Besides, the running time increases with longer prediction horizons under the same target cruise speed.

    To obtain an intuitive view of the safety performance, the values of the safety barrier keep positive in dense traffic scenarios, indicating that the EV's position always stays within the safe obstacle-free region $\mathcal{S}$. Moreover, in a cruise scenario with the target cruise speed $v_d = 15\,\text{m/s}$, the ST-RHC with prediction horizon $T = 8\,\text{s}$ can proactively avoid the first vehicle with a distance $d_a =16.894\,\text{m}$. In contrast, the ST-RHC with $T = 5\,\text{s}$ and $T = 2\,\text{s}$ avoids the first vehicle with a smaller distance of $13.661\,\text{m}$ and $9.925\,\text{m}$, respectively. Likewise, this phenomenon can be observed in the other two cruise scenarios, indicating that ST-RHC with a longer horizon has a better predictive ability for obstacle avoidance in dense traffic scenarios. As for cruise errors, increasing $T$ from $2\,\text{s}$ to $5\,\text{s}$ decreases the cruise MAE more than increasing $T$ from $5\,\text{s}$ to $8\,\text{s}$ when the target cruising speed is $10\,\text{m/s}$ and $20\,\text{m/s}$. Moreover, when $v_d = 15\,\text{m/s}$, the ST-RHC with a prediction horizon of $T = 5\,\text{s}$ achieves the lowest tracking MAE. These results show that increasing the predictive horizon has a limited effect on tracking performance.
    
    In terms of energy consumption, increasing $T$ from $2\,\text{s}$ to $5\,\text{s}$ leads to a significant decrease in mean absolute acceleration, by $54.18\,\%$, $32.72\,\%$, and $53.14\,\%$ for cruise speeds of $v_d = 10\,\text{m/s}$, $v_d = 12\,\text{m/s}$, and $v_d = 15\,\text{m/s}$, respectively. However, the ST-RHC with a prediction horizon of $T = 5\,\text{s}$ achieves the lowest mean absolute acceleration among the three different prediction horizons when the target cruising speed is $v_d = 10\,\text{m/s}$ or $20\,\text{m/s}$. This indicates that longer prediction horizons may not necessarily lead to better driving performance and that the choice of the prediction horizon should be based on specific scenarios to achieve better real-time performance. These results showcase a tradeoff between safety margin and tracking performance, which can be adjusted by varying the prediction horizon in our proposed ST-RHC framework. Besides, 
    the proposed ST-RHC enables the EV to safely track different cruise speeds in real time, which further demonstrates its generalizability.
     
\section{Conclusions}
	\label{sec:con} 
     This paper presents a novel computationally efficient ST-RHC scheme for safe and efficient autonomous driving in dense traffic. The ST-RHC considers both spatial and temporal relationships between the EV and surrounding HVs, enabling proactive collision avoidance in the presence of inaccurate prediction errors of HVs. 
     We thoroughly compare our ST-RHC to Batch-MPC and Frenet planner in various driving tasks and show that ST-RHC renders superior performance in both simulated and real-world datasets, with the advantage of our approach regarding safety, tracking accuracy, and optimization time. Besides, we performed ablation experiments to investigate the efficiency of the spatiotemporal safety barrier module in handling uncertain HVs. 
     Finally, we assessed the computational efficiency concerning varying collision avoidance constraints and evaluated the tradeoff between safety and task performance in tracking different goal speeds. 
    As part of our future work, the ST-RHC framework can be extended to address navigation problems under perception uncertainties in physical autonomous driving scenarios.
 
	\bibliographystyle{IEEEtran}
	\bibliography{egbib}
\end{document}